\documentclass[11pt]{article}
\usepackage{arxiv}

\usepackage{times}
\usepackage{latexsym}
\usepackage[T1]{fontenc}
\usepackage[utf8]{inputenc}
\usepackage{microtype}
\usepackage{inconsolata}
\usepackage{graphicx}
\usepackage{booktabs}
\usepackage{array}
\usepackage{xcolor}
\usepackage{amsmath}
\usepackage{amssymb}
\usepackage{paralist}
\usepackage{enumitem}
\usepackage{multirow}
\usepackage{pgfplots}
\pgfplotsset{compat=1.18}
\usepackage{tikz}
\usepackage{hyperref}
\hypersetup{colorlinks=true,allcolors={black!70!blue}}
\usepackage{natbib}

\usepackage{caption}
\title{A Two-Phase Stability Study of\\
LLM Judges and Bar Council Examiners\\
on Thai Bar-Exam Free-Form Essays}

\author{%
  Pawitsapak Akarajaradwong \\
  VISAI AI \\
  \texttt{pawitsapaka\_visai@vistec.ac.th}
  \And
  Wuttikrai Lertprasertphakorn \\
  Independent Researcher \\
  \texttt{msp.wuttikrai@gmail.com}
  \And
  Chompakorn Chaksangchaichot \\
  VISAI AI \\
  \texttt{chompakornc\_pro@vistec.ac.th}
  \And
  Sarana Nutanong \\
  Vidyasirimedhi Institute of Science and Technology \\
  \texttt{sarana.n@vistec.ac.th}
}

\begin{document}
\maketitle

\begin{abstract}
Free-form legal essay evaluation in NLP treats expert inter-rater stability as a single ceiling number, and treats LLM-judge agreement with that ceiling as evidence of judge stability. We test both assumptions on the Thai bar examination through an identical-inputs protocol: three Bar Council-trained examiners (A, B, C) and a 26-LLM judge panel score the same 15 cross-graded answers from the same four inputs (question, official Bar Council grading regulation, gold answer, candidate answer). The headline finding is asymmetric. On 10 of 15 cells where the rubric prescribes both axes, all 29 raters converge in a tight band: panel agreement is universal. On the remaining 5 cells where the rubric does not prescribe how to grade a correct final answer that omits a decisive statutory citation, the human panel splits between two coherent readings (B/C majority at the upper rubric band, score $6$--$8$; A minority at the lower band, score $1$--$2$). The LLM judge population does not split symmetrically: 22 of 26 LLMs score in or near B/C's contested band, 3 sit in the regulation-silent middle gap, and only 1 (GPT-5.4 Nano) approaches A's band without consistently scoring within it. \emph{Zero LLMs in our 26-judge panel reproduce the minority human reading on the contested cells.} The B/C-direction cluster spans every model size, vendor, and price tier we tested. An instrumented three-LLM anchor sub-panel (Claude 4.6 Opus, Gemini 3.1 Pro, GPT-5.4 Pro) carries determinism probes, input ablations, and bootstrap CIs, and reaches anchor panel $\alpha = 0.77$ on the 15 cells against human-panel $\alpha = 0.36$. The high LLM-panel $\alpha$ reflects systematic convergence on the majority reading rather than balanced reproduction of both readings; a benchmark that selects its LLM judge by maximising agreement with a human reference panel will inherit this asymmetry by construction.
\end{abstract}

\section{Introduction}
\label{sec:intro}

Open-ended generation in NLP is typically evaluated with n-gram overlap~\citep{papineni2002bleu,lin2004rouge}, LLM-as-a-judge protocols that grade win rate or per-facet quality~\citep{zheng2023judging,liu2023g,gu2024survey}, or task-tailored rubrics~\citep{es2024ragas,laban2024summary,akarajaradwong2025nitibench,magesh2025hallucinationfree}. The recurring assumption is that the panel agreement among credentialed human experts defines a single stability ceiling, and that an LLM judge has succeeded when its agreement against that panel approaches that ceiling~\citep{zheng2023judging,liu2023g,thakur2025judgebench,artstein2008inter}. Two empirical questions are rarely asked together: is the expert stability ceiling actually flat across items, and if not, does the LLM judge population reproduce both sides of the variance, or only one?

We test both halves of the assumption on the Thai bar examination. The bar exam is an attractive testbed because four quantities can be held identical across raters: the question, the official Bar Council grading regulation, the gold answer, and the participant's free-form essay. Three Bar Council-trained examiners (A, B, C) and a 26-LLM judge panel graded the same 15 cross-graded answers from those identical inputs.\footnote{The anonymised corpus (questions, gold answers, 150 essay answers, primary and cross-grade human scores, 1{,}164 span-anchored examiner comments) is publicly released at \url{https://huggingface.co/datasets/airesearch/thai-bar-exam-judging} under CC-BY 4.0.} Within the 26-LLM panel, an instrumented three-LLM anchor sub-panel (Claude 4.6 Opus, Gemini 3.1 Pro, GPT-5.4 Pro), chosen for vendor diversity and standardisation with prior NLP LLM-judge work, carries our determinism probes, input ablations, and bootstrap CIs at panel $\alpha = 0.77$. The remaining 23 LLMs extend the panel to test the structural claim at population scope. The human panel reaches Krippendorff $\alpha = 0.36$ on the same 15 cells.

Decomposing by question, the gap is entirely localised to one question (Q2) where the regulation does not prescribe how to grade a correct final answer that omits a decisive statutory citation. On the 10 cells from the other two questions, the human and LLM panels are stability-equivalent at $\alpha = 0.75$. On the 5 contested Q2 cells, one examiner's stricter reading of the regulation-silent boundary pulls the human panel to $\alpha = -0.34$. The interesting question is what the LLM judge population does on these contested cells.

The 26-LLM extension reveals an asymmetric structural finding. Classifying each judge by its contested-cell mean ($m_c$) on the 3 contested Q2 cells, $22$ of $26$ LLMs (85\%) land in or near B/C's contested reading band ($m_c \in [5, 8]$); $3$ LLMs sit in the regulation-silent middle gap ($3 < m_c < 5$); only $1$ LLM (GPT-5.4 Nano, $m_c = 2.33$) approaches A's reading band, and even Nano does not consistently score within A's band of $[1, 2]$. \emph{Zero LLMs in our 26-judge panel reproduce A's reading of the regulation-silent boundary.} The B/C-direction cluster reaches panel $\alpha = 0.56$ on the 15 cells; the single A-direction model achieves only $r(A, \text{Nano}) = 0.22$ across the 15-cell vector. The LLM population converges on the majority human reading and misses the minority human reading. The B/C-direction cluster spans every model size, vendor, and price tier in the panel; the closest-to-P LLM is a small closed-frontier model with low overall coherence.

The mechanism of this asymmetry is direct. Conventional LLM-as-a-judge protocols select a judge by maximising agreement with a human reference panel (typically the B/C-majority by aggregate). On rubric-silent items, this selection criterion lands on the B/C reading by construction: a judge that takes A's reading would have lower agreement with the B/C-skewed panel mean and would be filtered out. The high LLM-panel $\alpha$ that benchmark papers commonly cite measures how tightly LLM judges agree with each other, not whether the LLM panel covers the full space of legitimate human readings. On contested cells the two are different: 26 LLMs agree tightly with each other and tightly with B/C, but they collectively miss A entirely.

We adopt a two-phase design so that the role-separation pattern can be isolated. In Phase~1 (Section~\ref{sec:phase1}), eight LLMs sampled across closed-frontier large and open-weight tiers sit the same three Civil and Commercial Code essay questions as 42 Thai law students under fully-blinded examiner-graded conditions: the examiners did not know LLMs were in the participant pool at all. The LLM cohort mean ($13.12$/$30$) sits inside the student distribution. In Phase~2 (Sections~\ref{sec:phase2-anchor}--\ref{sec:phase2-clusters}), eight of the 26 LLM judges are the same models that sat the exam in Phase~1. Treating the two roles as one experimental design lets us state, quantitatively, that examinee capability rank in Phase~1 does not predict judge cluster alignment in Phase~2 (Section~\ref{sec:dualrole}): weaker Phase-1 examinees become more strongly B/C-aligned as judges, not less. Scaling capability does not push the LLM judge population toward the minority reading; if anything, it pushes further away.

Free-form essay rubrics in interpretation-intensive domains combine a deterministic correctness axis (grounded by the gold answer) with an interpretive quality axis (judged without one). Expert panels are stability-equivalent to LLM panels on items where both axes are rubric-determined. On items where the interpretive axis is regulation-silent, the human panel splits into multiple coherent readings and the LLM judge population reproduces only the majority. We call these regulation-silent items the rubric's \emph{interpretive bottleneck}; the asymmetric LLM convergence on them is what the rest of this report examines.

\paragraph{Contributions.} In order of conceptual weight:
\begin{compactenum}
    \item \textbf{Population-level asymmetric convergence finding under an identical-inputs two-phase protocol.} Across 26 LLM judges spanning closed-frontier large/small, flagship open-weight, and small open-weight tiers, no LLM reproduces the minority human reading on regulation-silent rubric items in the Thai bar examination. $22$ of $26$ LLMs converge on the majority (B/C) reading band on contested cells; $3$ sit in the regulation-silent gap; $1$ approaches the minority (A) reading without consistently scoring within it. The asymmetry holds across model size, vendor, and price. The two-phase design re-uses the same 8 LLMs as Phase-1 examinees and Phase-2 judges, so capability in one role is empirically separable from stability in the other; capability does not move judges toward the minority reading (Section~\ref{sec:dualrole}).
    \item \textbf{Three-LLM anchor reference panel with full instrumentation.} Within the 26-LLM panel, three closed-frontier judges (Claude 4.6 Opus, Gemini 3.1 Pro, GPT-5.4 Pro) carry determinism probes, input ablations, and bootstrap CIs at panel $\alpha = 0.77$ on the 15 cross-graded cells, against human-panel $\alpha = 0.36$. The anchor sub-panel is the configuration adopted by downstream benchmarks that grade free-form Thai legal answers under the same Bar Council rubric.
\end{compactenum}

\section{Background: The Thai Bar Examination}
\label{sec:background}

The Thai bar examination is the national qualifying assessment for legal practice in Thailand and a prerequisite for appointment to the judicial bench. It is administered twice a year by the Institute of Legal Education of the Thai Bar Council under the Ministry of Justice. A candidate sits four examination sessions, one per substantive doctrinal area (Civil and Commercial Code; Criminal Code; Civil Procedure Code; Criminal Procedure Code); each session presents ten free-form long-form questions, each scored 0--10 by a Bar Council-certified examiner, and a session pass requires roughly half marks. The bar is the entry point to a regulated profession; questions are constructed by sitting examiners, often current or former judges and senior practitioners, and graded by examiner panels selected from the same pool. The grading instrument is a published regulation that ranks free-form essay answers on an 8-tier ordinal scale.

This section makes three points the rest of the paper depends on. First, the bar exam is a free-form long-form essay assessment, not a multiple-choice instrument; the entire panel-agreement question we study below is meaningless on closed-form items. Second, the grading regulation has a specific structural shape (two crossed axes with overlapping tier-to-score ranges) that admits one regulation-silent boundary where competing legitimate interpretations are possible; that boundary is where the empirical stability fault line lands. Third, examiners do not undergo per-session calibration before grading; they apply the regulation under their own training and prior bar-grading experience, which is the realistic measurement setting any legal-NLP benchmark inherits when it cites a single credentialed expert as ground truth.

\subsection{Grading regulation and tier structure}
\label{sec:reg-overview}

The official grading instrument is the regulation of the Executive Committee of the Office of Legal Education Training of the Thai Bar Council. Appendix~\ref{app:regulation} reproduces the faithful translation; the load-bearing structure is summarised here. The regulation specifies an 8-tier ordinal scale crossing two axes: final-answer correctness against the gold answer (matches / differs), and reasoning quality (unusable / some / acceptable / good / very good). Table~\ref{tab:rubric-bg} gives the tier-to-score map.

\begin{table}[t]
\centering
\small
\setlength{\tabcolsep}{4pt}
\begin{tabular}{l c c c c c}
\toprule
& \multicolumn{5}{c}{\textbf{Reasoning quality}} \\
\cmidrule(lr){2-6}
\textbf{Final answer} & Unusable & Some & Acceptable & Good & Very good \\
\midrule
\emph{matches gold}    & T4: $0$--$1$  & T5: $2$--$5$  & T6: $5$--$6$  & T7: $7$--$8$ & T8: $9$--$10$ \\
\emph{differs}         & T1: $0$       & T2: $1$--$2$  & --            & T3: $2$--$4$ & -- \\
\bottomrule
\end{tabular}
\caption{The 8-tier rubric crosses final-answer correctness against reasoning quality. Grader picks a tier, then a half-integer score in its range. Ranges overlap and break monotonicity (T4 caps at $1$, below T3 at $2$--$4$); the regulation-silent T4-vs-T7 boundary on a correct answer with omitted decisive citation drives the disagreement in Section~\ref{sec:phase2-anchor}.}
\label{tab:rubric-bg}
\end{table}

Three properties of this rubric matter empirically. First, the numeric score alone does not recover the tier: a score of $1$ could be T2 (incorrect, somewhat acceptable reasoning) or T4 (correct, unusable reasoning); a score of $5$ could be T5 or T6. Our LLM judges return both a tier number and a score within that tier's range, mirroring the human two-step procedure (Appendix~\ref{app:judgeprompt}). Second, the regulation overlaps and inverts: a correct final answer with unusable reasoning (T4: $0$--$1$) scores strictly below an incorrect final answer with good reasoning (T3: $2$--$4$), so the rubric is not a clean correctness split. Third, the regulation prescribes tier categories but not the natural-language anchors that distinguish ``somewhat acceptable'' from ``acceptable'' reasoning. Examiners fill this calibration from their training and prior experience. The stability ceiling we measure in Section~\ref{sec:phase2-anchor} is on this shared-experience-plus-regulation calibration, not on the rubric axes in isolation.

\subsection{The deterministic and interpretive axes}
\label{sec:axes}

We label the two axes \emph{deterministic correctness} (final-answer-matches-gold against the official gold answer issued by the examiner committee) and \emph{interpretive quality} (the reasoning-quality scale, judged without a gold target). The deterministic axis is grounded by an external artefact: the gold answer fixes which side of the matches-or-differs split the candidate sits on. The interpretive axis is grounded by examiner judgement under the same regulation; the regulation gives the tier names and score ranges, but does not specify how to map a particular fact-by-fact reasoning chain into ``unusable / some / acceptable / good / very good'' on the correct-answer side.

For most fact patterns, the interpretive axis is rubric-determined in practice: the gold answer names the decisive Code sections, and examiners agree that an answer omitting a decisive section sits in the lower tiers and an answer naming all decisive sections sits in the upper tiers. For a smaller set of fact patterns, the interpretive axis is regulation-silent: a candidate reaches the correct final conclusion via a reasoning chain that names some but not all of the gold answer's decisive sections, and the regulation does not specify whether the omission is a T4 (correct, unusable) or T7 (correct, good) phenomenon. Section~\ref{sec:phase2-anchor} shows that the human stability ceiling fractures specifically on this regulation-silent boundary, and Section~\ref{sec:phase2-clusters} traces how the LLM judge population responds on the same boundary. The structural framing is that free-form essay rubrics combine a deterministic axis and an interpretive axis, and panel agreement degrades where the interpretive axis is regulation-silent.

\subsection{Examiner qualification and contamination control}
\label{sec:examiners-bg}

The three examiners (Judge A, Judge B, Judge C) are practising Thai legal professionals certified to grade at the Bar Council level. Each is a sitting Bar Council-trained examiner with prior bar-grading experience under the same regulation. The three examiners authored the three Q1--Q3 questions and the corresponding gold answers \emph{before} any participant answer was collected. The instrument is therefore not an item that any LLM could have encountered online before the 2026 session, which rules out training-data contamination on the questions and gold answers by construction. The Code itself and the grading regulation are public documents; we address residual contamination risk on doctrinal patterns in Section~\ref{sec:limitations} and via a gold-withheld input ablation in Section~\ref{sec:determinism}.

Bar Council scheduling and examiner availability impose constraints we accept and document rather than fight. Each examiner primary-graded all 50 answers for one question (A on Q1, B on Q2, C on Q3); the cross-grading subset (Section~\ref{sec:phase2-anchor}) is fully covered by all three examiners on each of its 15 cells. We did not run a pre-calibration round between the three examiners on a pilot subset before the live grading. Pre-calibrating examiners on each other would erase the interpretive variance that is the paper's main finding, because the realistic measurement setting any legal-NLP benchmark inherits when it cites a single credentialed expert as ground truth is the pre-calibration setting, not the post-calibration one.

\subsection{Doctrinal scope and the three questions}
\label{sec:questions-bg}

The three questions are drawn from the Civil and Commercial Code session and span three doctrinal clusters that bar candidates routinely encounter: ordinary-partnership liability (Q1, CCC Sections 1012, 1025, 1033, 1050); company director authority and good-faith third parties (Q2, CCC Sections 1022, 1023, $1023/1$); and shareholder annulment of a procedurally irregular general-meeting resolution (Q3, CCC Sections 1178, 1193, 1195). Each question is worth 10 points and a conventional pass on a three-question session is 15. Q1 turns on distinguishing the partner-internal management rule (Section 1033) from the partner-external liability rule (Section 1050); Q2 turns on the good-faith exception under Section $1023/1$ to the general no-bind rule of Section 1023; Q3 turns on the one-month limitation on annulment of unlawful resolutions under Section 1195 versus the unbounded standing to challenge fabricated minutes. Appendix~\ref{app:questions} reproduces all three questions and their gold answers in faithful English translation.

The regulation-silent boundary that drives the Phase~2 stability fault line lands on Q2. The three contested Q2 cells are those where the candidate reaches the correct final answer on sub-question (a) but argues from Section 1023 alone, without naming Section $1023/1$ as the decisive good-faith exception. Section~\ref{sec:qualitative} reproduces the verbatim translated candidate answer for Item~C1, the canonical instance, and the corresponding human comments and LLM-judge justifications.

\section{Related Work}
\label{sec:related}

\paragraph{Legal NLP and bar-style evaluation.} \citet{katz2024gpt4bar} report GPT-4 exceeding the U.S. Uniform Bar Examination passing threshold, a result dominated by multiple-choice components. \citet{guha2023legalbench} assemble 162 common-law tasks across legal reasoning categories, and \citet{fei2023lawbench} extend the pattern to Chinese law; both rely primarily on closed-form metrics or single-reference free-form scoring. \citet{chalkidis2022lexglue} and \citet{niklaus2023lextreme} provide multi-task multi-lingual legal benchmarks but stop short of free-form essay grading. In Thai law, \citet{akarajaradwong2025nitibench} benchmark retrieval-augmented question answering against two automatic scorers (Coverage and Contradiction under a GPT-4o judge); none of these report three-way human cross-grading on identical inputs, and none ask whether LLM judges are more or less stable than expert panels on free-form legal essays. \citet{magesh2025hallucinationfree} document hallucination in production legal-research tools but evaluate factual reliability, not panel stability.

\paragraph{LLM-as-a-judge.} \citet{zheng2023judging} establish the LLM-as-a-judge paradigm on open-ended dialogue with MT-Bench and Chatbot Arena, reporting GPT-4-vs-human agreement approaching human-vs-human agreement on chatbot responses. Most follow-ups implicitly target high single-human-to-LLM correlation as the success criterion. \citet{liu2023g} extend the paradigm to NLG quality with chain-of-thought-conditioned prompting (G-Eval). \citet{thakur2025judgebench} evaluate the judges themselves, asking when LLM judges fail to align with multi-human aggregates. \citet{wang2023judging} note that LLM evaluators are not fair in pairwise comparison settings. The wider survey \citep{gu2024survey} consolidates the field. A persistent finding line documents systematic biases in LLM judges: positional bias, verbosity bias, and self-preference effects \citep{zheng2023judging,koo2023benchmarking,panickssery2024llm}. These biases motivate our conservative configuration (fixed input order, single forward pass at temperature $0$, no chain-of-thought scaffolding, no few-shot examples, no tool use, no retrieval) and our preservation of full judge justifications so bias audits can be reproduced (Appendix~\ref{app:judgeprompt}). Our contribution is orthogonal to the bias line: we ask not whether an LLM judge agrees with one human, but whether multiple LLM judges agree with each other more than three human experts do, on identical inputs, and where in the rubric the disagreements lie.

\paragraph{Expert-panel consistency on subjective decisions.} Large-scale reviewer- and grader-consistency studies repeatedly document that expert verdicts on subjective tasks carry substantial panel-identity variance. The 2014 NeurIPS experiment routed 166 papers to two independent program committees and found that roughly half of the papers accepted by one were rejected by the other; the 2021 replication reproduces the same disagreement on a five-times-larger submission pool, with \citet{cortes2021inconsistency} estimating that roughly half of the variation in reviewer quality scores is subjective in origin. Writing-assessment practice \citep{shaw2007ielts} and the automated-essay-scoring literature \citep{ke2019automated} report similar patterns on high-stakes essay grading: trained rater panels do not converge to a single true score, and agreement is raised through training and adjudication rather than existing a priori. The structural mechanism we identify (panel agreement holds where the rubric is specified, diverges where it is silent) is consistent with these findings.

\paragraph{Inter-rater stability and the value of disagreement.} Krippendorff's $\alpha$ \citep{krippendorff2011computing} and Cohen's weighted $\kappa$ are the standard panel-level inter-rater stability metrics for ordinal scoring, with \citet{artstein2008inter} the canonical computational-linguistics framing. Recent NLP annotation scholarship \citep{plank2022problem} argues that legitimate disagreement among qualified annotators on subjective tasks should be modelled as informative signal rather than treated as noise to average out. Our two-cluster diagnosis on the regulation-silent boundary sits in this tradition: the three certified examiners disagree not because any is mis-calibrated but because the regulation leaves interpretive latitude at one specific tier boundary (T4 vs.\ T7, ``unusable'' vs.\ ``good'' reasoning when a decisive statutory citation is omitted on a correct final answer). What we add is the identical-inputs LLM-judge counterpart on the same items, a 29-rater cluster analysis that recovers the same fault line at population level without using vendor or size labels, and a per-question decomposition that localises the disagreement to a specific rubric zone rather than treating the panel as uniformly subjective.

\paragraph{Thai legal NLP.} Thai-language legal NLP remains sparse. \citet{akarajaradwong2025nitibench} provides a Thai Civil and Commercial Code retrieval-augmented question-answering benchmark, and \citet{akarajaradwong2026fourcorners} releases a production knowledge graph unifying Thai statutes, regulations, and court decisions; neither covers free-form essays or expert-graded panel stability. To our knowledge this is the first three-way human and 26-LLM judge stability study on identical-input Thai legal essays, the first to insert closed-frontier LLM answers into a Thai bar-style expert grading queue under full blinding, and the first to assemble a Thai bar-essay corpus with all six certified raters' scores and full justifications attached.

\section{Study Design}
\label{sec:study}

The study was run as a bar-qualification-style mock examination in early 2026. It is a two-phase design under identical inputs at each phase: Phase~1 puts LLMs in the examinee seat under blinded examiner grading; Phase~2 puts LLMs in the judge seat reading the same essays from the same four inputs every other rater receives. The two phases share a corpus by construction: the same 150 answers are produced in Phase~1 and graded in Phase~2. The full 15-cell cross-graded score matrix used in the Phase-2 stability analysis appears as Table~\ref{tab:humans} in Section~\ref{sec:phase2-anchor}.

\subsection{Corpus and instrument}

Fifty participants were enrolled: 42 Thai law students and legal practitioners preparing for the bar examination under informed consent, and 8 LLMs (Claude 4.6 Opus, Gemini 3.1 Pro, GPT-5.4 Pro, DeepSeek v3.2 Speciale, Qwen 3.5 Plus, GLM-5, Minimax M2.5, Grok 4.1 Fast) inserted into the participant roster as sequence identifiers indistinguishable from the human slots. Each participant produced one answer per question on the three Civil and Commercial Code questions of Section~\ref{sec:questions-bg}, yielding $50 \times 3 = 150$ primary-graded answers. Five answers (participant IDs 4, 6, 10, 21, 27; one of which is Claude 4.6 Opus, still unidentified to the cross-graders) were drawn for cross-grading by the other two examiners, giving $5 \times 3 = 15$ cross-graded cells. Section~\ref{sec:phase2-anchor} adds the LLM-judge axis on the same 15 cells.

The grading instrument is the Bar Council regulation summarised in Table~\ref{tab:rubric-bg} (verbatim translation in Appendix~\ref{app:regulation}). The three examiners authored the questions and the gold answers before any participant answer was collected; the instrument is therefore not on the open web at the time of the 2026 session, which removes question and gold-answer contamination on the LLM side by construction. The 2014 grading regulation and the Civil and Commercial Code itself are public.

\subsection{Phase 1: examinee sitting and primary grading}
\label{sec:study-phase1}

Human participants sat the session proctored, with a three-hour budget, no electronic aids, writing longhand on the standard Bar Council answer booklet. Each LLM was queried once per question through its provider API at temperature $0$ with a zero-shot system prompt instructing it to produce flowing Thai legal prose in the register of a handwritten student answer, with no retrieval, no tools, and no in-context examples. The LLM-system-prompt language is reproduced verbatim in Appendix~\ref{app:judgeprompt}; the prompt was held constant across all eight LLMs. LLM outputs were transcribed into the same answer-booklet format used by the human participants, all booklets were pseudonymised to sequence-identifier participant IDs, and the booklets were shuffled before grading. The three examiners were not informed that any participant was an LLM; the blinding was full, not partial.

Each examiner primary-graded one full question across all 50 participants under the regulation in Table~\ref{tab:rubric-bg}. Judge A primary-graded all 50 answers for Q1, Judge B for Q2, Judge C for Q3. Each primary score is one tier number ($1$--$8$) and one half-integer score in the tier's range. Examiners returned hand-written grading comments alongside the score on every answer they primary-graded. The Phase-1 score table is reported in Section~\ref{sec:phase1}.

\subsection{Phase 2: cross-grading and LLM-judge scoring}
\label{sec:study-phase2}

For each of the 15 cross-graded cells, the two examiners who had not primary-graded that question independently scored the answer given the question text, the official gold answer, and the grading regulation. The two cross-graders never saw the primary grader's score, the other cross-grader's score, or the participant's identity beyond the sequence ID.

On the same 15 cells, an LLM-judge panel grades from identical four-tuple inputs (question text, regulation, gold answer, participant answer). The three-LLM anchor sub-panel comprises Claude 4.6 Opus, Gemini 3.1 Pro, and GPT-5.4 Pro; the extended 26-LLM panel in Section~\ref{sec:phase2-clusters} adds 23 further models spanning closed frontier large and small, flagship open-weight, and small open-weight, sampled through OpenRouter at temperature $0$. Each LLM judge receives, in a single forward pass at temperature $0$, the question text, the grading regulation reproduced verbatim in the system prompt, the official gold answer, and the participant answer. The judge returns a strict JSON object with a tier number ($1$--$8$), a half-integer score within that tier's range, and a Thai-language justification (Appendix~\ref{app:judgeprompt}). No chain-of-thought scaffolding, no tools, no retrieval, and no few-shot examples. The judge prompt is identical across all $29$ LLM judges. Eight of the $26$ extended-panel judges are the same model that sat the exam in Phase~1; the dual-role analysis in Section~\ref{sec:dualrole} exploits this overlap.

Human cross-graders and LLM judges thus received the same four inputs and had no access to the primary grader's score, to each other's scores, to per-judge aggregates, or to the participant's identity. The stability comparison in the rest of the paper is therefore a comparison of scoring functions on identical information.

We deliberately did not pre-calibrate the three examiners against each other on a pilot subset before grading. The stability ceiling we measure is the one any legal-NLP benchmark inherits when it cites a single credentialed expert as ground truth without prior cross-rater calibration; pre-tuning would erase the interpretive variance that is the paper's main finding.

\subsection{Determinism, robustness, and ablations}

For each of the three anchor LLM judges, we additionally re-score the 15 cross-graded cells three times at temperature $0$ with seeds $44$, $45$, $46$, yielding $135$ additional judgements that bound the intra-judge stability under the same prompt. Two LLM-disadvantaged input ablations on the three-judge anchor panel measure dependence on the regulation and the gold answer respectively: in the regulation-withheld configuration, the judge sees only the gold answer (the tier definitions are excised from the system prompt); in the gold-withheld configuration, the judge sees only the regulation. Both ablations are reported in Section~\ref{sec:determinism}.

\section{Phase 1: LLMs as Examinees}
\label{sec:phase1}

The Phase~2 panel-stability comparison in Sections~\ref{sec:phase2-anchor}--\ref{sec:phase2-clusters} is meaningful only if the 150 answers are genuinely hard to judge: obvious rubbish or obvious perfection would render expert disagreement uninteresting. Phase~1 produces the answer corpus by sitting both 42 Thai law students and 8 LLMs through the same three Civil and Commercial Code essay questions under blinded examiner-graded conditions. This section reports the resulting score distribution. Three observations matter for the rest of the paper. First, the LLM cohort sits inside the student distribution, not above it; the grading task is not saturated. Second, the per-question pattern differs substantially across LLMs and across questions, so capability rank from Phase~1 carries information that the role-separation finding in Section~\ref{sec:dualrole} can test against. Third, the same 8 LLM identifiers reappear in the Phase-2 judge panel; the dual-role finding is intelligible only because Phase~1 fixes a per-LLM examinee score.

\subsection{Score distribution against the human cohort}

Table~\ref{tab:phase1-perq} reports per-question and total scores for all 8 LLMs alongside the human cohort means. The 8-LLM mean total ($13.12$/$30$) exceeds the 42-student mean ($12.04$) by about one point, well within per-question variance (LLM SD $5.28$ on totals; student SD $4.65$). Claude 4.6 Opus and Gemini 3.1 Pro tie at $20$/$30$, one point below the top human ($21$); five of the eight LLMs sit at or below the human median ($11.5$); none clears $21$. The LLM distribution overlaps the human distribution rather than dominating it.

\begin{table}[t]
\centering
\small
\setlength{\tabcolsep}{4pt}
\begin{tabular}{lrrrr}
\toprule
\textbf{Examinee} & \textbf{Q1} & \textbf{Q2} & \textbf{Q3} & \textbf{Total} \\
\midrule
Claude 4.6 Opus           & 6.0 & 7.0 & 7.0 & 20.0 \\
Gemini 3.1 Pro            & 9.0 & 6.0 & 5.0 & 20.0 \\
GPT-5.4 Pro               & 2.0 & 7.0 & 7.0 & 16.0 \\
DeepSeek v3.2 Speciale    & 2.0 & 4.0 & 6.0 & 12.0 \\
Qwen 3.5 Plus             & 0.0 & 7.0 & 4.0 & 11.0 \\
GLM-5                     & 0.0 & 3.0 & 6.0 &  9.0 \\
Minimax M2.5              & 1.0 & 5.0 & 3.0 &  9.0 \\
Grok 4.1 Fast             & 4.0 & 3.0 & 1.0 &  8.0 \\
\midrule
LLM mean ($n=8$)          & 3.00 & 5.25 & 4.88 & 13.12 \\
Human mean ($n=42$)       & 3.86 & 4.45 & 3.73 & 12.04 \\
$\Delta$ (LLM $-$ human)  & $-0.86$ & $+0.80$ & $+1.14$ & $+1.08$ \\
\bottomrule
\end{tabular}
\caption{Phase~1 primary-graded scores for the 8 LLM examinees and the 42-student human cohort means. Each score is the assigned examiner's tier-and-score under the rubric in Table~\ref{tab:rubric-bg}. The LLM cohort sits inside the human distribution; the per-question pattern shows a clean inversion (LLMs lose on Q1, gain on Q2 and Q3).}
\label{tab:phase1-perq}
\end{table}

\begin{figure}[t]
\centering
\begin{tikzpicture}
\definecolor{cClaude}{HTML}{D95F02}
\definecolor{cGemini}{HTML}{1B9E77}
\definecolor{cGPT}{HTML}{7570B3}
\definecolor{cDeepSeek}{HTML}{E7298A}
\definecolor{cQwen}{HTML}{66A61E}
\definecolor{cMinimax}{HTML}{E6AB02}
\definecolor{cGLM}{HTML}{A6761D}
\definecolor{cGrok}{HTML}{555555}
\begin{axis}[
    width=0.95\textwidth,
    height=4.5cm,
    xmin=0, xmax=30,
    xtick={0,5,10,15,20,25,30},
    xlabel={Total score (out of 30)},
    xlabel style={font=\footnotesize, yshift=2pt},
    tick label style={font=\scriptsize},
    ymin=-0.15, ymax=1.2,
    hide y axis,
    axis x line*=bottom,
    tick style={color=black!50},
    clip=false,
    legend style={
        at={(0.5,-0.45)},
        anchor=north,
        font=\tiny,
        legend columns=4,
        column sep=3pt,
        row sep=-2pt,
        draw=black!25,
        fill=white,
        inner xsep=3pt, inner ysep=1.5pt,
    },
    legend image post style={scale=0.8},
]
\fill[green!8] (axis cs:15, 0) rectangle (axis cs:30, 1.1);
\draw[dashed, gray!55] (axis cs:15, 0) -- (axis cs:15, 1.1);
\node[gray!70, font=\tiny, anchor=south west] at (axis cs:15.2, 1.0) {pass $\geq 15$};
\addplot[only marks, mark=*, mark size=1.3pt, mark options={fill=black!28, draw=black!50, fill opacity=0.75}] coordinates {
    (5,0.30) (5,0.75) (6,0.50) (6.5,0.50)
    (7,0.28) (7,0.55) (7,0.82)
    (8,0.10) (8,0.28) (8,0.45) (8,0.60) (8,0.78) (8,0.95)
    (9,0.15) (9,0.32) (9,0.50) (9,0.68) (9,0.85)
    (10,0.50) (11,0.32) (11,0.72)
    (12,0.18) (12,0.40) (12,0.60) (12,0.85)
    (14,0.30) (14,0.72)
    (15,0.22) (15,0.42) (15,0.65) (15,0.88)
    (16,0.32) (16,0.72) (17,0.50)
    (18,0.28) (18,0.50) (18,0.75)
    (19,0.25) (19,0.50) (19,0.78)
    (20,0.50) (21,0.50)
};
\addlegendentry{Humans ($n{=}42$)}
\addplot[only marks, mark=square*, mark size=3pt, mark options={fill=cClaude, draw=black!40}] coordinates {(20,0.32)};
\addlegendentry{Claude 4.6 Opus (20)}
\addplot[only marks, mark=triangle*, mark size=3.5pt, mark options={fill=cGemini, draw=black!40}] coordinates {(20,0.70)};
\addlegendentry{Gemini 3.1 Pro (20)}
\addplot[only marks, mark=diamond*, mark size=3.5pt, mark options={fill=cGPT, draw=black!40}] coordinates {(16,0.50)};
\addlegendentry{GPT-5.4 Pro (16)}
\addplot[only marks, mark=pentagon*, mark size=3.5pt, mark options={fill=cDeepSeek, draw=black!40}] coordinates {(12,0.50)};
\addlegendentry{DeepSeek v3.2 Sp.\ (12)}
\addplot[only marks, mark=square*, mark size=3pt, mark options={fill=cQwen, draw=black!40}] coordinates {(11,0.50)};
\addlegendentry{Qwen 3.5 Plus (11)}
\addplot[only marks, mark=triangle*, mark size=3.5pt, mark options={fill=cMinimax, draw=black!40}] coordinates {(9,0.30)};
\addlegendentry{Minimax M2.5 (9)}
\addplot[only marks, mark=diamond*, mark size=3.5pt, mark options={fill=cGLM, draw=black!40}] coordinates {(9,0.70)};
\addlegendentry{GLM-5 (9)}
\addplot[only marks, mark=pentagon*, mark size=3.5pt, mark options={fill=cGrok, draw=black!40}] coordinates {(8,0.50)};
\addlegendentry{Grok 4.1 Fast (8)}
\draw[blue!70!black, dotted, line width=0.7pt] (axis cs:12.04, 0) -- (axis cs:12.04, 1.1);
\draw[red!70!black, dotted, line width=0.7pt] (axis cs:13.12, 0) -- (axis cs:13.12, 1.1);
\end{axis}
\end{tikzpicture}
\caption{Phase~1 total scores by participant. Humans ($n = 42$, gray dots) and 8 LLMs inserted blind (colored shapes) share a single jittered band; shading marks the conventional $\geq 15$ pass zone; dotted lines are the human ($12.04$, blue) and LLM ($13.12$, red) cohort means.}
\label{fig:phase1_scores}
\end{figure}
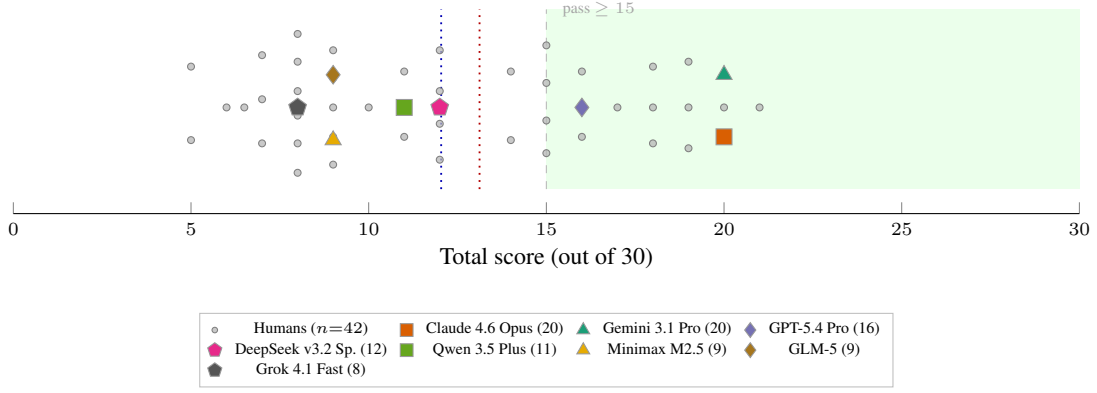

\subsection{Per-question pattern}

The aggregate near-parity masks a clean per-question structure. Q1 is the hardest item for the LLM cohort: the decisive move is to distinguish partner-internal management (CCC Sections 1025, 1033) from third-party-external liability (Sections 1012, 1050) and to apply the latter to the fact pattern. Models that cite only one side of the internal/external distinction lose most of their credit. Three of eight LLMs score zero or one on Q1; only Gemini 3.1 Pro reaches the upper band ($9$). Q2 and Q3 reward more mechanical doctrinal pattern-matching once the candidate has reached the correct final answer, and the LLM mean matches or exceeds the human cohort mean on both. This is consistent with the broader Phase-2 finding (Section~\ref{sec:phase2-anchor}) that the LLM judge population converges easily on the deterministic axis of the rubric and disagrees on the interpretive axis: the same models that converge on the deterministic axis as judges also outscore humans on the more mechanical questions as examinees.

\subsection{Spread within the LLM set}

The 12-point spread among the 8 LLMs ($8$ to $20$) is wide: a frontier-label branding does not map cleanly onto Thai-legal-essay capability. Claude 4.6 Opus and Gemini 3.1 Pro at $20$ tie the second-best human (at $21$/$30$); GPT-5.4 Pro at $16$ is mid-band; DeepSeek v3.2 Speciale, Qwen 3.5 Plus, GLM-5, and Minimax M2.5 ($9$--$12$) cluster with median students; Grok 4.1 Fast at $8$ sits above only the four weakest students. The Phase-1 capability rank is anchored: the same eight models reappear as judges in Section~\ref{sec:phase2-clusters}, and Section~\ref{sec:dualrole} uses this rank to test whether examinee capability predicts judge cluster alignment. The short answer is that it does not: the weakest examinee (Grok 4.1 Fast, $8$/$30$) is the strongest in-cluster judge by the B/C-cluster strength metric, and the inverse correlation between examinee total and judge cluster strength is exploratory ($n = 8$) but consistently signed.

The bar is the entry point to a regulated profession; the conventional pass on a three-question session is $15$/$30$. Two of eight LLM examinees clear that threshold ($25\%$), against $15$ of $42$ students ($35.7\%$). Whether one phrases this as ``LLMs are at the median student level on Thai bar essays'' or ``LLMs underperform the pass rate of human candidates'' depends on framing; the data permits both. The point that matters for the stability question is that Phase~1 produced a corpus where no answer is trivially gradable in either direction, and the 15 cross-graded cells we use in Phase~2 (Section~\ref{sec:phase2-anchor}) span the full $0$--$10$ range under the assigned primary grader.

\section{Phase 2: Instrumented Three-LLM Anchor Sub-Panel}
\label{sec:phase2-anchor}

This section reports the Phase-2 result on the instrumented three-LLM anchor sub-panel (Claude 4.6 Opus, Gemini 3.1 Pro, GPT-5.4 Pro) against the three-human-examiner panel (Judges A, B, C) on the 15 cross-graded cells, with all six raters reading the same four inputs (question, regulation, gold answer, participant answer). The three LLMs were chosen at study start for vendor diversity and standardisation with prior NLP LLM-judge work, before any score was collected; they are not the highest-mean-$r$-with-humans models in our 26-LLM extension (Section~\ref{sec:phase2-clusters}). They are the sub-panel we ran deeper measurement on: three-seed determinism probe at $T = 0$ (Section~\ref{sec:determinism}), two input ablations (regulation-withheld, gold-withheld), bootstrap CIs at $5{,}000$ resamples (Appendix~\ref{app:bootstrap}), and pair-by-pair $\kappa_w$ and $\rho$. The 26-LLM extension in Section~\ref{sec:phase2-clusters} tests the structural claim at population scope on the same 15 cells.

On the 15 cross-graded cells, panel-level Krippendorff $\alpha$ is $0.36$ for the human panel and $0.77$ for the three-LLM anchor sub-panel. The full per-cell score matrix and per-rater means are reported in Table~\ref{tab:humans}; the headline column means are $4.67$, $6.33$, $6.30$ for A, B, C and $5.93$, $6.13$, $6.20$ for Claude, Gemini, GPT-5.4 Pro. Within-cell panel standard deviation is $1.32$ for humans and $0.71$ for LLMs, a $1.87\times$ ratio.

\begin{table}[t]
\centering
\small
\setlength{\tabcolsep}{4pt}
\begin{tabular}{l|ccc|ccc}
\toprule
\textbf{Cell} & \textbf{A} & \textbf{B} & \textbf{C} & \textbf{Cla} & \textbf{Gem} & \textbf{GPT} \\
\midrule
C1$^\dagger$  & 1.0 & 7.0 & 8.0 & 5.5 & 6.0 & 5.5 \\
C2$^\dagger$  & 1.0 & 7.0 & 7.0 & 7.0 & 5.5 & 7.0 \\
C3$^\dagger$  & 2.0 & 6.0 & 6.5 & 5.5 & 6.0 & 5.5 \\
C4            & 0.0 & 4.0 & 3.0 & 4.0 & 4.0 & 5.5 \\
C5            & 5.0 & 6.0 & 6.0 & 7.5 & 9.0 & 8.0 \\
C6            & 9.0 & 9.0 & 9.0 & 9.0 & 10.0 & 9.0 \\
C7            & 8.0 & 8.0 & 8.0 & 7.0 & 8.0 & 7.5 \\
C8            & 3.0 & 4.0 & 5.0 & 4.0 & 4.0 & 5.0 \\
C9            & 3.0 & 4.0 & 6.0 & 4.0 & 4.0 & 5.0 \\
C10           & 6.0 & 6.0 & 7.0 & 7.0 & 4.0 & 6.0 \\
C11           & 5.0 & 7.0 & 5.0 & 5.5 & 6.0 & 4.5 \\
C12           & 5.0 & 7.0 & 6.0 & 4.5 & 6.0 & 5.5 \\
C13           & 6.0 & 6.0 & 5.0 & 4.0 & 5.5 & 5.5 \\
C14           & 8.0 & 7.0 & 7.0 & 7.5 & 6.0 & 6.0 \\
C15           & 8.0 & 7.0 & 6.0 & 7.0 & 8.0 & 7.5 \\
\midrule
mean          & 4.67 & 6.33 & 6.30 & 5.93 & 6.13 & 6.20 \\
SD            & 2.90 & 1.45 & 1.49 & 1.61 & 1.88 & 1.29 \\
panel $\alpha$ & \multicolumn{3}{c|}{\textbf{0.36}} & \multicolumn{3}{c}{\textbf{0.77}} \\
\bottomrule
\end{tabular}
\caption{All 15 cross-graded cells scored by the three human examiners (A, B, C) and the three anchor LLM judges (Claude 4.6 Opus, Gemini 3.1 Pro, GPT-5.4 Pro) under the same four inputs (question, regulation, gold answer, candidate answer). $^\dagger$ marks the three contested Q2 cells (C1, C2, C3) that drive the human-panel collapse: A scores $1$, $1$, $2$ where B and C score $6$--$8$. The three LLM anchors score in the middle range ($5.5$--$7$) on these contested cells, close to B/C's reading. Panel $\alpha$ across all 15 cells: humans $0.36$, three-LLM anchor sub-panel $0.77$.}
\label{tab:humans}
\end{table}

\subsection{The gap is localised to one question}
\label{sec:localisation}

Table~\ref{tab:localisation} decomposes panel stability by question.

\begin{table}[t]
\centering
\small
\setlength{\tabcolsep}{5pt}
\begin{tabular}{lcccc}
\toprule
\textbf{Slice} & $N$ & \textbf{Human $\alpha$} & \textbf{LLM $\alpha$} & \textbf{Gap} \\
\midrule
Q1                & 5  & $0.78$  & $0.79$  & $+0.01$ \\
Q2                & 5  & $-0.34$ & $0.68$  & $\mathbf{+1.02}$ \\
Q3                & 5  & $0.55$  & $0.67$  & $+0.12$ \\
\midrule
Q1 and Q3 pooled  & 10 & $\mathbf{0.75}$ & $\mathbf{0.75}$ & $0.00$ \\
All Questions     & 15 & $0.36$  & $0.77$  & $+0.41$ \\
\bottomrule
\end{tabular}
\caption{Panel-level Krippendorff $\alpha$ decomposed by question. Q1 and Q3 give equivalent human and LLM stability. The cross-panel gap concentrates on 5 Q2 cells.}
\label{tab:localisation}
\end{table}

On Q1 and Q3 the two panels are equivalently stable ($\alpha = 0.75$ each on the 10-cell pool). The full-panel gap of $+0.41$ is driven by the 5 Q2 cells, three of which (Items C1, C2, C3) are the cells where the candidate reaches the correct final answer but omits a decisive statutory citation (Section $1023/1$). On those contested cells the human panel falls to $\alpha = -0.34$ while the LLM panel holds at $\alpha = 0.68$. The gap is a property of one regulation-silent boundary, not of base-rate panel disagreement on bar-essay scoring.

\subsection{The Q2 contested cells}

The rubric (Table~\ref{tab:rubric-bg}) crosses final-answer correctness against reasoning quality. The deterministic axis is grounded by the gold answer; the interpretive axis is judged without one. The score range for correct-answer-with-unusable-reasoning (T4) is $0$--$1$ and for correct-answer-with-good-reasoning (T7) is $7$--$8$; the regulation does not specify which side an answer that omits a decisive citation falls on. All three human cross-graders agree on correctness for the contested cells but split on the reasoning category. On Item~C1, A scores $1$ (\emph{unusable}); B and C score $7$ and $8$ (\emph{good}); the three LLM judges give $5.5$/$6$/$5.5$ (\emph{acceptable}). Every LLM judge names Section $1023/1$ verbatim in its justification (9 of 9 judge-cell combinations on the contested Q2 cells): all six raters see the same fact and disagree on its category. Section~\ref{sec:qualitative} reproduces the verbatim translated candidate answer and all six justifications for Item~C1.

\subsection{Locating the variance with a mixed panel}
\label{sec:mixed-panel}

Pairwise Pearson on the 15 cross-graded cells gives B--C at $r = 0.75$ and the three LLM judges at $r = 0.74$--$0.84$, with B and C pairing with Claude and Gemini at $r$ in $[0.55, 0.76]$. Judge A sits outside this cluster: mean $r = 0.57$ against the LLMs, $0.52$ against the other two humans. B, C, Claude, and Gemini all sit inside a single cross-tribe cluster; A's mean cross-tribe Pearson sits below the mean within-tribe value of every other grader, reflecting A's stricter interpretive reading on the T4-vs-T7 boundary.

A one-rater-swap sweep on all 20 three-rater panels (Table~\ref{tab:panels}) confirms that the variance is not a property of any specific triple. Every panel including A lands at $\alpha \leq 0.49$; every panel excluding A lands at $\alpha \geq 0.63$. Replacing A with any one of $\{B, C, \text{Cla}, \text{Gem}, \text{GPT}\}$ produces a panel in the upper group. The variance is a single legitimate interpretive split on the regulation-silent cells, not a property of one specific examiner or one specific model.

\begin{table}[t]
\centering
\small
\setlength{\tabcolsep}{3pt}
\begin{tabular}{lc@{\hspace{1.2em}}lc}
\toprule
\multicolumn{2}{c}{\textbf{With A ($n=10$)}} & \multicolumn{2}{c}{\textbf{Without A ($n=10$)}} \\
\textbf{Panel} & $\alpha$ & \textbf{Panel} & $\alpha$ \\
\midrule
A, C, GPT        & 0.30 & C, Gem, GPT        & 0.63 \\
A, B, GPT        & 0.35 & B, C, GPT          & 0.65 \\
A, B, C          & 0.36 & C, Cla, Gem        & 0.65 \\
A, C, Gem        & 0.38 & B, C, Gem          & 0.67 \\
A, C, Cla        & 0.39 & C, Cla, GPT        & 0.69 \\
A, Cla, GPT      & 0.42 & B, Cla, GPT        & 0.71 \\
A, B, Cla        & 0.44 & B, Gem, GPT        & 0.72 \\
A, Gem, GPT      & 0.44 & B, C, Cla          & \textbf{0.72} \\
A, B, Gem        & 0.46 & B, Cla, Gem        & 0.73 \\
A, Cla, Gem      & 0.49 & Cla, Gem, GPT      & \textbf{0.77} \\
\midrule
\textbf{max}     & \textbf{0.49} & \textbf{min}        & \textbf{0.63} \\
\bottomrule
\end{tabular}
\caption{Panel $\alpha$ for all 20 three-rater combinations from $\{A, B, C, \text{Cla}, \text{Gem}, \text{GPT}\}$, sorted within each column by $\alpha$. Every panel including A sits at $\alpha \leq 0.49$; every panel excluding A sits at $\alpha \geq 0.63$. Removing A from every panel it appears in raises $\alpha$ by at least $0.14$.}
\label{tab:panels}
\end{table}

\subsection{Pairwise coefficients}

Cohen $\kappa_w$ (integer-rounded scores on the $0$--$10$ ordinal scale) and Spearman $\rho$ on the 15 cells give within-human means of $0.47$ and $0.52$ against within-LLM means of $0.71$ and $0.73$. The within-LLM mean is $1.5\times$ the within-human mean under both coefficients, smaller than the $2.1\times$ panel-$\alpha$ gap but in the same direction. Pairwise averaging dilutes any single-rater outlier effect, which is why panel $\alpha$ widens the ratio relative to pairwise coefficients. The full pair-by-pair table is in Appendix~\ref{app:bootstrap}.

\subsection{Bootstrap and significance}

We bootstrap-resample the 15 cells with replacement $5{,}000$ times and recompute panel $\alpha$ on each resample. The Krippendorff $\alpha$ for the human panel is $0.36$ with $95\%$ CI $[-0.09, 0.71]$; the three-LLM anchor sub-panel is $0.77$ with $95\%$ CI $[0.54, 0.91]$. The point estimate of $\Delta\alpha$ (LLM $-$ human) is $+0.39$ with $95\%$ CI $[0.03, 0.76]$; LLM $-$ human is positive in $98.5\%$ of resamples. Swapping A for Claude in a mixed panel (B, C, Cla) reaches $\alpha = 0.72$ with $95\%$ CI $[0.41, 0.86]$ and the swap-A-to-Claude $\Delta\alpha$ is positive in $98.8\%$ of resamples. Both direction contrasts have positive lower bounds. The bootstrap table is reproduced in Appendix~\ref{app:bootstrap}.

\subsection{Leave-one-out checks}

Two leave-one-out sweeps check that the contrast is not carried by a single rater or a single item. Dropping one rater at a time, the human-pair $\alpha$ ranges from $0.22$ (drop B) to $0.76$ (drop A); the LLM-pair $\alpha$ ranges from $0.73$ (drop GPT) to $0.80$ (drop Gemini). The A-less pair $(B, C)$ at $\alpha = 0.76$ sits inside the LLM range; the every-LLM-dropped range $[0.73, 0.80]$ rules out the inversion being carried by any specific model. Dropping one cross-graded cell at a time, the human range is $[0.23, 0.49]$ and the LLM range is $[0.67, 0.82]$; the two ranges are disjoint. Excluding only the three contested Q2 cells (Items C1, C2, C3), the remaining 12 cells give human $\alpha = 0.73$ and LLM $\alpha = 0.79$: the panels are stability-equivalent on the non-contested subset. The whole-panel gap is a property of three specific items at the regulation-silent boundary, not of base-rate disagreement on bar-essay scoring. The full leave-one-out tables are in Appendix~\ref{app:bootstrap}.

\section{Asymmetric LLM Convergence on the Contested Boundary}
\label{sec:phase2-clusters}

The anchor sub-panel of Section~\ref{sec:phase2-anchor} establishes that a small instrumented LLM-judge panel agrees more than the human panel, and that the gap is localised to one regulation-silent rubric boundary. Two questions remain at the population level. First, is the LLM-panel-above-human-panel direction a property of just those three anchor LLMs, or does it replicate across a wider LLM judge population? Second, does the wider LLM population split along the same fault line as the human panel, reproducing both the B/C-majority reading and the A-minority reading, or does it converge on one reading and miss the other? This section addresses both by extending the judge panel to 26 LLMs sampled across closed-frontier large and small, flagship open-weight, and small open-weight tiers, each scoring the same 15 cells from the same four inputs.

The headline empirical finding is asymmetric. The LLM-population-above-human-panel direction replicates: 22 of 26 LLMs (85\%) score in or near the B/C reading band on the contested cells; 3 LLMs (12\%) sit in the regulation-silent middle gap; only 1 LLM (GPT-5.4 Nano, $m_c = 2.33$) approaches the A reading band, and even Nano does not consistently score within A's contested-cell band of $[1, 2]$. \emph{No LLM in the 26-judge panel reproduces A's reading of the regulation-silent boundary.} This is the structural finding of the section. We treat it as a population-level pattern of the LLM judge ensemble on this rubric, not as a single-model failure mode.

\subsection{Extended panel}

The 26-LLM panel adds 23 models to the three anchors: Claude 4.7 Opus, Claude Sonnet 4.6, Claude Haiku 4.5, GPT-5.5 Pro, GPT-5.4 Mini, GPT-5.4 Nano, Gemini 3 Flash, Gemini 3.1 Flash Lite, Grok 4.1 Fast, Kimi K2.6, Qwen 3.6 Max, Qwen 3.6 Plus, Qwen 3.6 Flash, Qwen 3.6 35B-A3B, Qwen 3.5 Plus, GLM 5, GLM 5.1, Gemma 4 26B-A4B, Gemma 4 31B, DeepSeek v4 Pro, DeepSeek v4 Flash, DeepSeek v3.2 Speciale, Minimax M2.5. Each judge is queried at temperature $0$ with the same prompt as the three anchors (Appendix~\ref{app:judgeprompt}), routed through OpenRouter. The judge returns a tier number, a half-integer score, and a Thai-language justification per cell, and the judge has access to no other information. The 29 raters in scope from this point are the three human cross-graders, the three anchor LLM judges, and the 23 extended-panel LLMs.

\subsection{Contested-cell direction: where each judge lands on the fault line}
\label{sec:two-clusters}

\begin{figure}[t]
\centering
\includegraphics[width=\columnwidth]{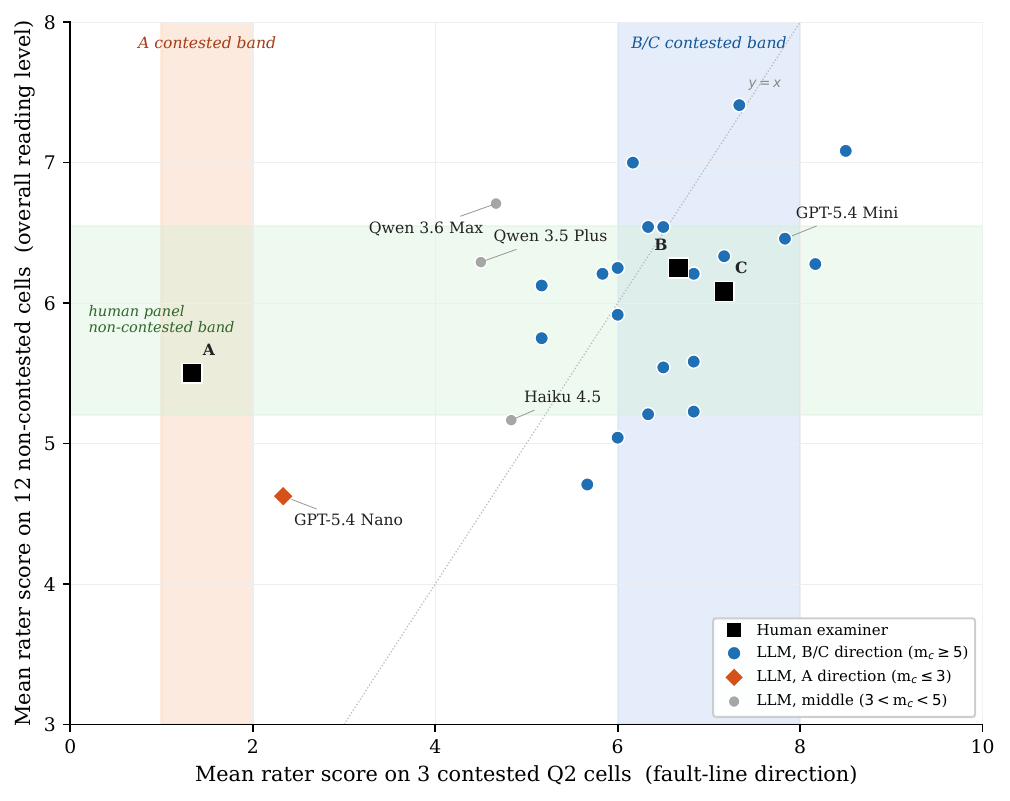}
\caption{Each rater positioned by mean score on the 3 contested Q2 cells ($x$, fault-line direction) and the 12 non-contested cells ($y$, overall reading level). Vertical bands mark the human-panel reading bands on contested cells: A at $[1, 2]$ (orange) and B/C at $[6, 8]$ (blue). The horizontal band marks the human-panel non-contested range. All 29 raters fall inside the horizontal band: agreement on the 12 non-contested cells is universal. On the contested cells, the human panel splits between A's band (1 examiner) and B/C's band (2 examiners), but the LLM judge population does not split symmetrically. $22$ of $26$ LLMs (blue circles) land in B/C-direction ($m_c \geq 5$); $3$ LLMs (gray circles) sit in the regulation-silent middle gap; only $1$ LLM (GPT-5.4 Nano, orange diamond) approaches A's band, without consistently scoring within it. Zero LLMs reproduce A's reading. Marker colour reflects contested-cell direction; no vendor or size label is used.}
\label{fig:clusters}
\end{figure}

Figure~\ref{fig:clusters} plots each rater on the contested-vs-non-contested cell-mean plane. The horizontal axis is the rater's mean score on the 3 contested Q2 cells (fault-line direction). The vertical axis is the rater's mean score on the 12 non-contested cells (overall reading level). The vertical reference bands are the human-panel reading bands on the contested cells: A's band at $[1, 2]$ (orange) and B/C's band at $[6, 8]$ (blue). The horizontal reference band is the human-panel non-contested range ($[5.5, 6.3]$). All 29 raters fall inside the horizontal band: panel agreement on the 12 non-contested cells is universal. The horizontal spread along the fault-line axis is the structural object of interest.

We classify each judge by its contested-cell mean using a distance-to-band rule. A judge is in \emph{A-direction} if its contested mean is at most $3.0$ (within $1$ unit of A's band); in \emph{B/C-direction} if its contested mean is at least $5.0$ (within $1$ unit of B/C's band); in the \emph{middle} otherwise. The thresholds are pre-specified from the human-panel reading bands and apply equally to every rater. Under this rule:

\begin{itemize}[itemsep=0pt,topsep=2pt]
    \item \textbf{B/C-direction LLMs ($n = 22$).} Panel $\alpha = 0.556$ on the 15 cells. With B and C added, the panel $\alpha = 0.549$ on $n = 24$.
    \item \textbf{Middle LLMs ($n = 3$).} Qwen 3.6 Max ($m_c = 4.67$), Qwen 3.5 Plus ($m_c = 4.50$), Haiku 4.5 ($m_c = 4.83$). Panel $\alpha = 0.540$ on these three. They sit in the regulation-silent gap between the two reading bands.
    \item \textbf{A-direction LLMs ($n = 1$).} GPT-5.4 Nano alone, with $m_c = 2.33$. Nano is the closest LLM to A's reading band but does not consistently score within it. The pair Pearson $r(A, \text{Nano}) = 0.223$ on 15 cells, equivalent to a 2-rater $\alpha = 0.229$.
\end{itemize}

\paragraph{Why contested-cell direction rather than 15-cell correlation.}
A natural alternative classification uses 15-cell Pearson correlation against each human (the canonical inter-rater agreement measure). Under that rule with the same $r \geq 0.5$ threshold per human, Qwen 3.6 Max and Qwen 3.5 Plus would be ``A-aligned'' because their overall $r_A$ exceeds $0.5$. However, both of these judges score in the middle gap on the contested cells specifically ($m_c = 4.67$ and $4.50$); their high overall $r_A$ comes from agreement with A on the 12 non-contested cells where every rater agrees, not from siding with A on the fault line. Contested-cell direction is the more direct measure of fault-line reading.

\subsection{Per-rater table}

Table~\ref{tab:cluster-table} reports every LLM judge's contested-cell mean, non-contested-cell mean, and pairwise Pearson correlation with each of the three human cross-graders. Rows are grouped by contested-cell direction and sorted within each group by $m_c$.

\begin{table}[t]
\centering
\small
\setlength{\tabcolsep}{3pt}
\begin{tabular}{lrrrrr}
\toprule
\textbf{Judge} & $m_c$ & $m_{nc}$ & $r_B$ & $r_C$ & $r_A$ \\
\midrule
\multicolumn{6}{l}{\emph{A-direction ($m_c \leq 3$); $n = 1$, $\alpha(	ext{A, Nano}) = 0.23$}} \\
GPT-5.4 Nano               & 2.33 & 4.62 & $-0.03$ & $0.10$ & $0.22$ \\
\midrule
\multicolumn{6}{l}{\emph{Middle ($3 < m_c < 5$); $n = 3$}} \\
Qwen 3.5 Plus              & 4.50 & 6.29 & $0.32$ & $0.44$ & $0.67$ \\
Qwen 3.6 Max               & 4.67 & 6.71 & $0.37$ & $0.36$ & $0.64$ \\
Claude Haiku 4.5           & 4.83 & 5.17 & $0.29$ & $0.35$ & $0.49$ \\
\midrule
\multicolumn{6}{l}{\emph{B/C-direction ($m_c \geq 5$); $n = 22$; panel $\alpha = 0.56$}} \\
Kimi K2.6                  & 5.17 & 6.12 & $0.57$ & $0.77$ & $0.75$ \\
Gemma 4 31B                & 5.17 & 5.75 & $0.67$ & $0.56$ & $0.51$ \\
DeepSeek v3.2 Speciale     & 5.67 & 4.71 & $0.35$ & $0.39$ & $0.15$ \\
Gemini 3.1 Pro             & 5.83 & 6.21 & $0.76$ & $0.55$ & $0.61$ \\
Claude 4.6 Opus            & 6.00 & 5.92 & $0.74$ & $0.72$ & $0.59$ \\
GPT-5.4 Pro                & 6.00 & 6.25 & $0.61$ & $0.57$ & $0.52$ \\
GLM 5.1                    & 6.00 & 6.25 & $0.46$ & $0.59$ & $0.43$ \\
Qwen 3.6 Flash             & 6.00 & 5.04 & $0.51$ & $0.54$ & $0.25$ \\
Gemini 3 Flash             & 6.17 & 7.00 & $0.77$ & $0.74$ & $0.69$ \\
DeepSeek v4 Pro            & 6.33 & 5.21 & $0.37$ & $0.69$ & $0.16$ \\
Qwen 3.6 Plus              & 6.33 & 6.54 & $0.51$ & $0.54$ & $0.44$ \\
Grok 4.1 Fast              & 6.50 & 5.54 & $0.48$ & $0.75$ & $0.25$ \\
Gemma 4 26B-A4B            & 6.50 & 6.54 & $0.73$ & $0.67$ & $0.56$ \\
Claude 4.7 Opus            & 6.83 & 6.21 & $0.70$ & $0.68$ & $0.32$ \\
Claude Sonnet 4.6          & 6.83 & 5.58 & $0.60$ & $0.77$ & $0.38$ \\
Qwen 3.6 35B-A3B           & 6.83 & 5.23 & $0.42$ & $0.47$ & $0.00$ \\
DeepSeek v4 Flash          & 6.83 & 6.21 & $0.42$ & $0.38$ & $0.34$ \\
GLM 5                      & 7.17 & 6.33 & $0.61$ & $0.67$ & $0.38$ \\
GPT-5.5 Pro                & 7.33 & 7.41 & $0.83$ & $0.69$ & $0.58$ \\
GPT-5.4 Mini               & 7.83 & 6.46 & $-0.21$ & $0.28$ & $-0.28$ \\
Minimax M2.5               & 8.17 & 6.28 & $0.43$ & $0.32$ & $0.07$ \\
Gemini 3.1 Flash Lite      & 8.50 & 7.08 & $0.67$ & $0.67$ & $0.31$ \\
\midrule
\multicolumn{6}{l}{\emph{Human cross-graders (for reference)}} \\
A                          & 1.33 & 5.50 & $0.61$ & $0.43$ & $1.00$ \\
B                          & 6.67 & 6.25 & $1.00$ & $0.75$ & $0.61$ \\
C                          & 7.17 & 6.08 & $0.75$ & $1.00$ & $0.43$ \\
\bottomrule
\end{tabular}
\caption{Each rater's mean score on the 3 contested cells ($m_c$), mean score on the 12 non-contested cells ($m_{nc}$), and pairwise Pearson with each of the three human cross-graders on the full 15-cell vector. Rows grouped by contested-cell direction and sorted within group by $m_c$. The A-direction cluster has one member; no LLM scores within A's $[1, 2]$ contested band consistently. The B/C-direction cluster (22 of 26 LLMs) spans every model size, vendor, and price tier in the panel.}
\label{tab:cluster-table}
\end{table}

The asymmetry is severe. The B/C-direction cluster spans every model size, vendor, and price tier in the panel: closed frontier large (Opus 4.6/4.7, GPT-5.4 Pro, GPT-5.5 Pro, Gemini 3.1 Pro), closed frontier small (Sonnet 4.6, Gemini 3 Flash, Gemini 3.1 Flash Lite, GPT-5.4 Mini), flagship open-weight (Kimi K2.6, Qwen 3.6 Plus, GLM 5, DeepSeek v4 Pro), and small open-weight (Qwen 3.6 Flash, Gemma 4 26B-A4B, Gemma 4 31B). The A-direction cluster is a single small closed-frontier model (Nano). Even within Nano's behaviour, the agreement with A on 15 cells is only $r_A = 0.22$: Nano scores $1.0$ and $0.5$ on two contested cells (consistent with A's reading) but $5.5$ on the third (against A's reading), so the agreement is partial and unstable. By contrast, the 22 B/C-direction LLMs reach panel $\alpha = 0.56$ on the same 15 cells, an order-of-magnitude tighter coherence.

\subsection{Capability rank against direction}
\label{sec:capability-rank}

Cluster membership does not track size, vendor, or price. The B/C-direction cluster covers the full range of the panel; the single A-direction model (Nano) is a closed-frontier-small OpenAI model that fails on multiple other axes (Section~\ref{sec:failmodes}) and is not a representative ``A-aligned LLM judge'' that a benchmark could safely select.

A second view by mean inter-rater correlation across the 29-rater matrix gives an even sharper picture. The top-ranked rater by mean correlation with the other 28 is Claude 4.6 Opus ($\bar{r} = 0.73$); the second is the open-weight Gemma 4 26B-A4B ($\bar{r} = 0.71$), ahead of GPT-5.4 Pro, GPT-5.5 Pro, and Gemini 3.1 Pro. The three certified human examiners rank $22$, $24$, and $27$ of $29$ by the same metric (B at $\bar{r} = 0.51$; C at $\bar{r} = 0.55$; A at $\bar{r} = 0.39$). On a 26B-parameter open-weight model, the same scoring function reaches inter-rater consensus higher than every other rater except the largest closed-frontier model in the panel.

The phrase ``LLM judges are X capable'' carries no Phase-2 reading-direction commitment. The B/C-direction pattern is independent of size, vendor, or price; the systematic absence of A-direction LLMs is also independent of these axes. The next section diagnoses three specific judges (Haiku, Mini, Nano) that fail in qualitatively distinct ways on the rubric's two-axis structure; these three sit outside the B/C-direction cluster but the failure modes are not the same as ``reads like P.''

\section{Failure-Mode Diagnostics}
\label{sec:failmodes}

Three LLM judges in our 26-judge panel exhibit qualitatively distinct failure modes that each locate one corner of the rubric in Table~\ref{tab:rubric-bg}. Each failure mode lands in a different cell of the contested-cell-direction partition of Section~\ref{sec:two-clusters}. Claude Haiku 4.5 ($m_c = 4.83$) sits in the regulation-silent middle gap because it quantises the score scale to a pass/fail bucket. GPT-5.4 Nano ($m_c = 2.33$) is the single A-direction LLM in the panel; its strict-reasoning behaviour on contested cells reproduces the A reading, although its overall 15-cell correlation with A is only $r_A = 0.22$ because its scores on non-contested cells are noisy. GPT-5.4 Mini ($m_c = 7.83$) sits in B/C-direction by mean but anti-correlates with B at $r_B = -0.21$ because it scores leniently and uniformly high regardless of cell, producing high contested mean but low cell-to-cell variation match. This section reports the worked failure modes. The diagnostic tone here is intentional: we locate where each judge's variance sits on the two rubric axes; we do not propose to remove the judges from the panel.

\subsection{Scale quantization: Claude Haiku 4.5}
\label{sec:fm-haiku}

Claude Haiku 4.5 quantises the half-integer scale to two values. Every one of its 15 cross-graded scores is either $3.5$ or $7.5$ (Table~\ref{tab:haiku-quant}). No intermediate score appears. The full 150-cell baseline mean is $5.03$, sitting between the two quantised values, and the unique-value set on the baseline expands only by another two values (a handful of $1.5$ and $5.5$ instances on hard or easy cells; the bimodal $\{3.5, 7.5\}$ pattern dominates).

\begin{table}[t]
\centering
\small
\setlength{\tabcolsep}{4pt}
\begin{tabular}{r|ccc}
\toprule
\textbf{Participant} & \textbf{Q1} & \textbf{Q2} & \textbf{Q3} \\
\midrule
1   & 7.5 & 3.5 & 3.5 \\
2   & 3.5 & 3.5 & 7.5 \\
3   & 7.5 & 3.5 & 7.5 \\
4   & 7.5 & 3.5 & 3.5 \\
5   & 3.5 & 7.5 & 3.5 \\
\bottomrule
\end{tabular}
\caption{Claude Haiku 4.5 scores on all 15 cross-graded cells. Two unique values: $\{3.5, 7.5\}$.}
\label{tab:haiku-quant}
\end{table}

The Thai justifications produced by Haiku 4.5 are coherent legal analyses of the right length and shape; the model is engaging with the rubric. The mode collapse is not a reasoning collapse but a scale collapse: Haiku's score distribution sits on a pass/fail bucket and cannot express the mid-range scores B and C use on the contested cells (B's mean is $6.33$ with values across the full $4$--$10$ range; Haiku is silent in $4$--$7$ entirely). This is the rubric's bottom-of-T6 and bottom-of-T7 zone, where most of the cross-graded cells live.

\subsection{Final-answer-only grading: GPT-5.4 Mini}
\label{sec:fm-mini}

GPT-5.4 Mini grades the deterministic axis only. Its baseline mean ($6.39$) is the highest of any judge on the 150-cell baseline and sits $1.3$--$1.8$ points above the panel mean range of $4.6$--$5.1$ across the other 28 raters. On the contested Q2 cells where B and C give $6$--$8$, A gives $1$--$2$, and the three anchor LLM judges give $5.5$--$6$ in the middle, GPT-5.4 Mini gives $7.5$--$8.0$ (Table~\ref{tab:mini-final}). Mini sits above B and C rather than between B/C and A.

\begin{table}[t]
\centering
\small
\setlength{\tabcolsep}{4pt}
\begin{tabular}{r|cccccc}
\toprule
\textbf{Cell} & \textbf{A} & \textbf{B} & \textbf{C} & \textbf{Cla} & \textbf{Gem} & \textbf{Mini} \\
\midrule
C2     & 1.0 & 7.0 & 7.0 & 7.0 & 5.5 & 8.0 \\
C1     & 1.0 & 7.0 & 8.0 & 5.5 & 6.0 & 8.0 \\
C3     & 2.0 & 6.0 & 6.5 & 5.5 & 6.0 & 7.5 \\
\bottomrule
\end{tabular}
\caption{GPT-5.4 Mini on the three contested Q2 cells against A/B/C/Claude/Gemini.}
\label{tab:mini-final}
\end{table}

The Thai justifications produced by Mini consistently note that the candidate's reasoning is wrong or incomplete but the final conclusion is correct, and assign tier 7 (correct final answer, reasoning good) on that basis alone. The $r = -0.21$ Pearson with B is the signature of a lenient final-answer-only reading: Mini sees the deterministic axis (matches gold) and uses the upper-band tier-7/8 score range regardless of where the reasoning sits on the interpretive axis.

\subsection{Reasoning-strict grading: GPT-5.4 Nano}
\label{sec:fm-nano}

GPT-5.4 Nano grades the interpretive axis strictly and the deterministic axis as a soft constraint. On the contested Q2 cells, Nano sits in the regulation's $0$--$1$ band (T4, correct final answer with unusable reasoning), aligning with Judge A's stricter reading. On Item~C2 where B $=$ C $= 7$ and A $= 1$, Nano scores $1.0$; on Item~C3 where B $= 6$, C $= 6.5$, and A $= 2$, Nano scores $0.5$ (Table~\ref{tab:nano-strict}).

\begin{table}[t]
\centering
\small
\setlength{\tabcolsep}{4pt}
\begin{tabular}{r|cccccc}
\toprule
\textbf{Cell} & \textbf{A} & \textbf{B} & \textbf{C} & \textbf{Cla} & \textbf{Gem} & \textbf{Nano} \\
\midrule
C2     & 1.0 & 7.0 & 7.0 & 7.0 & 5.5 & 1.0 \\
C1     & 1.0 & 7.0 & 8.0 & 5.5 & 6.0 & 5.5 \\
C3     & 2.0 & 6.0 & 6.5 & 5.5 & 6.0 & 0.5 \\
\bottomrule
\end{tabular}
\caption{GPT-5.4 Nano on the three contested Q2 cells. Nano sits with A on Items C2 and C3; the three-LLM anchors sit in the middle.}
\label{tab:nano-strict}
\end{table}

The Thai justifications produced by Nano repeatedly state that the reasoning has major errors, name a specific omitted statutory citation, and drop the score below the regulation's mid-range under T4. This is the A reading expressed by an LLM: the candidate omitted a decisive citation, the omission decides the reasoning tier, and a correct final answer reached without the decisive citation belongs in T4 (correct final answer, unusable reasoning, $0$--$1$). The reading is internally consistent with one legitimate parsing of the regulation, and Nano commits to it on 2 of 3 contested cells (Item~C2 score $1.0$, Item~C3 score $0.5$). Nano is the only LLM in the 26-judge panel classified A-direction by contested-cell mean (Section~\ref{sec:two-clusters}), and even Nano's mean ($m_c = 2.33$) sits above A's actual band of $[1, 2]$ rather than within it. The strict-reasoning behaviour is real on contested cells but does not transfer to consistent agreement with A across the full 15-cell support, which is why Nano's overall 15-cell $r_A$ is only $0.22$.

\subsection{Three failures land on the two rubric axes}

The three failure patterns sit on the two axes Section~\ref{sec:axes} identified. Mini collapses onto the deterministic axis: final answer correct $\Rightarrow$ upper band, regardless of reasoning quality. Nano collapses onto the interpretive axis under the strict reading: reasoning omits decisive citation $\Rightarrow$ T4, regardless of final-answer correctness. Haiku coarsens the scale to a two-value bucket, so neither axis can drive a mid-range score. Below a capability threshold the judge collapses onto one axis or quantises the scale; above the threshold the judge converges on one of the two legitimate readings. The empirical cluster split (Section~\ref{sec:phase2-clusters}) is where the rubric's interpretive ambiguity lands at the population level once a judge does converge.

The distribution of these failure modes does not track vendor or branding: Haiku 4.5 is a closed-frontier-small Anthropic model, Mini and Nano are closed-frontier-small OpenAI models, and the three failure modes are not the same. Under contested-cell direction (Section~\ref{sec:two-clusters}), Haiku sits in the middle gap, Nano is the single A-direction LLM in the entire 26-judge panel, and Mini sits at the top of the B/C-direction band by mean but anti-correlates with B on cell-to-cell variation. Each failure pattern locates one corner of the two-axis rubric (scale coarsening; lenient final-answer-only; strict reasoning-only), and exactly one LLM in the panel (Nano) takes the minority human reading on the fault line. Descriptively, the failure-mode judges are the only judges in the panel that deviate from the majority B/C convergence, and Nano specifically is the only judge that deviates toward the minority reading rather than away from both.

\section{Dual-Role: Examinee vs.\ Judge}
\label{sec:dualrole}

Eight of the 26 LLM judges in Section~\ref{sec:phase2-clusters} are the same models that sat the exam in Phase~1 (Section~\ref{sec:phase1}). The dual-role overlap lets us test, directly and without confound, whether examinee capability predicts judge stability under the same scoring function on the same items. The short answer is that it does not: \emph{the same model is a different system as examinee and as judge.}

\subsection{Cluster alignment of the eight examinees as judges}

Table~\ref{tab:examinee_judge} reports each Phase-1 LLM examinee re-used as a Phase-2 judge against the three human cross-graders. Under contested-cell direction (Section~\ref{sec:two-clusters}), seven of the eight examinees take B/C direction on the fault line ($m_c \geq 5$); one (Qwen 3.5 Plus, $m_c = 4.50$) sits in the regulation-silent middle gap. \emph{Zero examinees take A direction.} The asymmetric LLM majority convergence of Section~\ref{sec:phase2-clusters} therefore replicates inside the Phase-1-examinee subset: even among LLMs that took the bar exam under blinded human grading, none reads the contested cells the way P does. The 8-rater LLM-examinee panel $\alpha = 0.57$ on the 15 cells, above the human panel $\alpha = 0.36$ but below the B/C-direction-cluster panel $\alpha = 0.56$ (which spans 22 LLMs). The reduced $\alpha$ here is consistent with the middle-gap position of Qwen 3.5 Plus rather than with a B/C-vs-A split inside the examinee panel.

\begin{table}[t]
\centering
\small
\setlength{\tabcolsep}{3pt}
\begin{tabular}{lc|cccc}
\toprule
\textbf{Examinee (as judge)} & \textbf{Exam} & $r_A$ & $r_B$ & $r_C$ & $\Delta$ \\
\midrule
Claude 4.6 Opus        & 20 & 0.59 & 0.74 & 0.72 & $+0.14$ \\
Gemini 3.1 Pro         & 20 & 0.61 & 0.76 & 0.55 & $+0.05$ \\
GPT-5.4 Pro            & 16 & 0.52 & 0.61 & 0.57 & $+0.07$ \\
DeepSeek v3.2 Speciale & 12 & 0.15 & 0.35 & 0.39 & $+0.22$ \\
Qwen 3.5 Plus          & 11 & 0.67 & 0.32 & 0.44 & $\mathbf{-0.29}$ \\
GLM-5                  &  9 & 0.38 & 0.61 & 0.67 & $+0.26$ \\
Minimax M2.5           &  9 & 0.07 & 0.43 & 0.32 & $+0.30$ \\
Grok 4.1 Fast          &  8 & 0.25 & 0.48 & 0.75 & $\mathbf{+0.36}$ \\
\bottomrule
\end{tabular}
\caption{Each Phase-1 LLM examinee scored as a Phase-2 judge on the 15 cross-graded cells. \textbf{Exam} is the examinee's Phase-1 bar-exam total out of $30$ (from Table~\ref{tab:phase1-perq}). $r_A$, $r_B$, $r_C$ are Pearson with the three human cross-graders. $\Delta = (r_B + r_C)/2 - r_A$ is positive when the judge clusters with the B/C majority reading and negative when the judge clusters with A.}
\label{tab:examinee_judge}
\end{table}

\subsection{Same model, different system}

The clean version of the role-separation claim uses the eight examinees as evidence that examinee skill does \emph{not} credential judge selection. If examinee skill transferred to judge stability, the three top-scoring Phase-1 examinees (Claude 4.6 Opus at $20$/$30$, Gemini 3.1 Pro at $20$/$30$, GPT-5.4 Pro at $16$/$30$) would also be the three strongest in-cluster judges by $\Delta$. They are not. They reach $\Delta = +0.14$, $+0.05$, $+0.07$ as judges, moderate-to-weak cluster strength. The strongest in-cluster judge across the 8-examinee panel is Grok 4.1 Fast, the \emph{lowest}-scoring Phase-1 examinee at $8$/$30$, which reaches $\Delta = +0.36$. The two next-strongest in-cluster judges are also low Phase-1 scorers: Minimax M2.5 ($\Delta = +0.30$ as judge; $9$/$30$ as examinee) and GLM-5 ($\Delta = +0.26$ as judge; $9$/$30$ as examinee). The heuristic ``select judges by examinee performance'' would have picked exactly the wrong models.

A direct comparison: Claude 4.6 Opus scores $20$/$30$ as an examinee and lands at B/C-direction $m_c = 6.00$ as a judge with $\Delta = +0.14$. Grok 4.1 Fast scores $8$/$30$ as an examinee, more than ten points lower, and lands at B/C-direction $m_c = 6.50$ as a judge with $\Delta = +0.36$, the strongest in-cluster pull in the eight-examinee panel. The two examinees sit at opposite ends of the Phase-1 capability rank and converge on the same reading as judges, one of them more strongly. The only examinee that deviates from the B/C direction at all is Qwen 3.5 Plus ($11$/$30$ in Phase~1), and even Qwen 3.5 Plus only reaches the middle gap ($m_c = 4.50$), not A direction. The Phase-1 rank does not predict the Phase-2 direction; what is independent of Phase-1 rank is the B/C-direction consensus that all 8 examinees approximately share.

The phrase ``the same model is a different system as examinee and as judge'' is loadbearing: \emph{the scoring function applied by Grok 4.1 Fast to a participant answer is empirically different from the scoring function Grok 4.1 Fast is itself scored under in Phase~1, even though the underlying model weights and inference configuration are identical.} What changes between roles is the input and the prompt: in Phase~1 the model produces a free-form Thai legal essay under no rubric prompt; in Phase~2 it reads the gold answer and the regulation and produces a tier-plus-score under the explicit rubric prompt. The same weights deliver different functions because the task framing is different. The dual-role design lets us state this empirically rather than as a conjecture.

\subsection{Inverse correlation, exploratory}

Across the eight examinees, Pearson(exam total, $\Delta$) $= -0.34$ over $n = 8$. Excluding the single anti-cluster outlier Qwen 3.5 Plus, the correlation rises to Pearson $= -0.90$ over $n = 7$. The exploratory reading is that a weaker examinee (under the regulation as a test-taker) clusters more strongly with B/C (under the regulation as a judge). We flag this finding as exploratory and report it for completeness; it is not statistically significant at $n = 8$, it depends on outlier exclusion, and it is post-hoc. A larger panel of LLM examinees would be needed to distinguish a true inverse correlation from an artefact of one outlier (Section~\ref{sec:limitations}).

The direction is consistent with a structural interpretation. If a weaker Phase-1 examinee has internalised the deterministic axis of the rubric (recognises a correct final answer when one is presented) but has not internalised the interpretive axis (cannot autonomously construct a high-quality reasoning chain), then as a Phase-2 judge the same model is graded on a task it can do (recognise the gold-answer-aligned final conclusion) rather than a task it cannot (construct the reasoning), and its score concentrates on the deterministic axis. The B/C-aligned reading on the contested cells is the more deterministic of the two readings: B and C credit a correct final answer with present-but-incomplete reasoning at the upper-mid band ($5.5$--$8$), where P withholds credit on the interpretive axis specifically. The inverse correlation, if real, says that the cleanest deterministic readers are the weaker examinees, not the stronger ones. We do not lean on this; it is a candidate explanation that future replication can test.

\subsection{Implication for evaluator selection}

The dual-role result has a consequence for any benchmark that selects an LLM judge by single-pass capability rank or by agreement-with-human-panel maximisation. First, scaling capability does not move judges toward the minority reading: the strongest Phase-1 examinees become moderate-strength B/C-direction judges; none becomes a A-direction judge. Second, the panel mean against which agreement is maximised is B/C-skewed by construction (B and C are 2 of 3 humans), so a judge that takes A's reading would have lower agreement with that mean and would be filtered out. The cluster composition of a selected LLM panel is therefore an empirical property of the candidate pool and the selection criterion together; further scaling alone does not change it.

\section{Qualitative Mechanism on the Contested Boundary}
\label{sec:qualitative}

This section reproduces one of the three contested Q2 cells in full so that the structural diagnosis of Section~\ref{sec:axes} can be read off a concrete answer and rubric application. We use Item~C1 because it carries the largest within-panel spread (A at $1$, B at $7$, C at $8$, the three anchor LLM judges clustered at $5.5$--$6$). The verbatim translated candidate answer is reproduced in Appendix~\ref{app:walkthrough}; the full set of human comments and LLM-judge justifications across three cells is in Appendix~\ref{app:justifications}. The summary below names what is on each side of the regulation-silent boundary in the candidate's own writing.

\subsection{Item C1: what the candidate does and does not do}

Item~C1 reaches the correct final answer on both sub-questions of Q2 (Section~\ref{sec:questions-bg}). On (a), the candidate concludes that Arctic Cooling Co.\ may demand the refrigerator price from Jipatha Electricals Co.; on (b), the candidate concludes that AI-Freeze Co.\ may demand damages from Jipatha Electricals Co. Both conclusions match the gold answer's bottom-line conclusions. The reasoning chain that supports each conclusion cites CCC Section 1023 (the general rule that a registered change of directors binds third parties only after publication in the Royal Gazette) and CCC Section 66 (the corporate-purpose rule); the chain does \emph{not} explicitly name CCC Section $1023/1$ (the good-faith exception to Section 1023, which the gold answer marks as the decisive provision for the (a) fact pattern).

The structural shape of the answer is therefore: correct final answer reached via a reasoning chain that names some but not all of the gold answer's decisive citations. The omission is on one specific provision the gold answer identifies as decisive. The candidate's surrounding analysis (good-faith reasoning under registration vs.\ publication, corporate purpose under Section 66) does engage the substance of Section $1023/1$ implicitly, but does not cite the section number. This is the regulation-silent boundary of Table~\ref{tab:rubric-bg}: the rubric's correct-final-answer side spans T4 ($0$--$1$, unusable reasoning) through T8 ($9$--$10$, very good reasoning), and the regulation does not specify how to grade a correct final answer that names the good-faith mechanism without naming the specific Section $1023/1$.

\subsection{Six raters, two readings}

Table~\ref{tab:c1-summary} reproduces the score and grading-comment summary for the six anchor raters on Item~C1.

\begin{table}[t]
\centering
\small
\setlength{\tabcolsep}{4pt}
\begin{tabular}{lc l}
\toprule
\textbf{Rater} & \textbf{Score} & \textbf{Reading of the omission} \\
\midrule
A (human, primary on Q1)   & 1.0 & Decisive. Minimal credit. \\
B (human, primary on Q2)   & 7.0 & Implicit but not named. Upper band. \\
C (human, primary on Q3)   & 8.0 & Acceptable. Upper band. \\
\midrule
Claude 4.6 Opus            & 5.5 & T6 (reasoning acceptable). \\
Gemini 3.1 Pro             & 6.0 & T6 (reasoning acceptable). \\
GPT-5.4 Pro                & 5.5 & T6 (reasoning acceptable). \\
\bottomrule
\end{tabular}
\caption{Six anchor raters on Item~C1. A, B, C are the three Bar Council human examiners; Claude, Gemini, GPT-5.4 Pro are the three anchor LLM judges. A's $1.0$ is at the bottom of T4; B's $7.0$ and C's $8.0$ are in T7; the three LLM judges are at the bottom of T6 / top of T5.}
\label{tab:c1-summary}
\end{table}

A, B, and C agree on the deterministic axis: the final answer is correct on both sub-questions. They split on the interpretive axis. A reads the absence of an explicit Section $1023/1$ citation as decisive: the candidate has not articulated the good-faith exception by name, so the reasoning is in T4 (correct final answer, unusable reasoning), and the score is $1$ at the top of that tier. B and C read the omission as a partial-credit element: the candidate's substantive analysis carries the load of Section $1023/1$ even though the section number is not named, so the reasoning is in T7 (correct final answer, reasoning good), and the score is $7$ or $8$ in that tier. The two readings are internally consistent with one legitimate parsing of the regulation each: T4 if a missing decisive citation is a tier-deciding fact, T7 if it is a within-tier fact. The regulation does not adjudicate between the two parsings.

The three anchor LLM judges sit at the bottom of T6: \emph{correct final answer, reasoning acceptable rather than good}, because the substantive Section $1023/1$ analysis is present but the section number is not named. Their per-cell justifications (Appendix~\ref{app:justifications}) read uniformly: the candidate is credited for the good-faith reasoning; the missing section number lowers the tier from T7 to T6 but does not eject the answer from the correct-final-answer side of the rubric. Every LLM judge names Section $1023/1$ verbatim in its own justification, identifying the citation the candidate did not name. The three LLM judges thus do not split between the B/C reading and the A reading on this cell; they take a third-route reading that lands closer to B and C than to A, which is consistent with their cluster membership in Section~\ref{sec:phase2-clusters}.

\subsection{What the qualitative case implies}

Two facts about the cell are diagnostic. First, all six raters see the same artefact: the LLM justifications and the human comments converge on which citation the candidate did not name. The disagreement is not on the facts of the candidate's answer; it is on the rubric category that an answer with this specific shape belongs in. Second, the disagreement is structured: A holds the strict reading on all three contested Q2 cells; B and C hold the upper-band reading on all three; the three anchor LLM judges hold the mid-band reading on all three. The fault line is not a single cell oddity; it is a population-level split that recurs whenever the regulation-silent boundary is engaged. Section~\ref{sec:phase2-clusters} shows the same fault line at 29-rater scale; this section shows the mechanism that produces it on one cell.

The structural implication is that the rubric, not the raters, is where the variance lives. An automated benchmark that nominates A's reading as ground truth gets a different LLM-judge $\alpha$ from one that nominates B/C's reading as ground truth, and both are consistent with the regulation. One consequence is that a single panel-$\alpha$ figure obscures which of the two coherent readings the panel converged on; partitioning judges by contested-cell direction (Section~\ref{sec:phase2-clusters}) recovers this information.

\section{Determinism and Robustness}
\label{sec:determinism}

The Phase-2 stability claim of Sections~\ref{sec:phase2-anchor}--\ref{sec:phase2-clusters} rests on the assumption that the LLM-judge function is reasonably stable under its own resampling. We check that assumption with a three-seed determinism probe at $T = 0$ and two LLM-disadvantaged input ablations on the three-LLM anchor sub-panel. The bootstrap and leave-one-out results from Section~\ref{sec:phase2-anchor} are summarised here in context; the full numerical tables live in Appendix~\ref{app:bootstrap}.

\subsection{Three-seed determinism probe at $T = 0$}

Each of the three anchor LLM judges (Claude 4.6 Opus, Gemini 3.1 Pro, GPT-5.4 Pro) was re-scored on the 15 cross-graded cells with three independent seeds at $T = 0$ (seeds $44$, $45$, $46$), yielding $135$ additional judgements. Within-judge cross-seed Krippendorff $\alpha$ is $0.998$ for Claude, $0.945$ for Gemini, and $0.877$ for GPT-5.4 Pro. Mean within-cell SD across seeds is $0.00$, $0.25$, and $0.39$ respectively, with a single GPT-5.4 Pro cell at SD $1.16$. Two facts are notable. First, $T = 0$ is not universally deterministic: only Claude is effectively deterministic, while Gemini and GPT-5.4 Pro carry non-trivial within-judge variance under nominally identical inputs. Second, the three-LLM anchor sub-panel $\alpha$ across the three independent $k$-slices stays in $[0.76, 0.82]$, just below the within-judge ceiling of $\approx 0.94$. Inter-model diversity reduces $\alpha$ below the determinism ceiling rather than inflating it. The headline $0.77$ panel-$\alpha$ is the diversity-suppressed value, not an artefact of identical-input-identical-output replay across judges, and the upper bound on what a single resample at $T = 0$ could produce is $\approx 0.94$.

\subsection{Input ablations: regulation-withheld and gold-withheld}

Two LLM-disadvantaged ablations test whether the LLM-panel agreement depends on having the grading regulation or the gold answer in the prompt. Humans retain both inputs throughout in all configurations.

\begin{table}[t]
\centering
\small
\setlength{\tabcolsep}{4pt}
\begin{tabular}{lc}
\toprule
\textbf{Configuration} & \textbf{LLM panel $\alpha$} \\
\midrule
Baseline (full inputs)         & $0.77$ \\
Regulation withheld            & $0.57$ \\
Gold answer withheld           & $0.47$ \\
\midrule
Human reference (full inputs)  & $0.36$ \\
\bottomrule
\end{tabular}
\caption{Three-LLM anchor sub-panel $\alpha$ under the two LLM-disadvantaged ablations. Both reduced configurations stay above the human reference.}
\label{tab:ablations}
\end{table}

Per-judge mean-score shifts under each ablation are substantial (GPT-5.4 Pro's mean shifts by $+0.87$ when the regulation is withheld; the maximum $|\Delta|$ on a single cell under gold-withheld is $3.5$), so the ablations genuinely perturb the judges rather than producing identical outputs under a different prompt. The panel direction (LLM $\alpha >$ human $\alpha$) survives both. Withholding the gold answer is the harsher of the two ablations: it forces each judge to construct the correct legal reasoning from the question alone before grading, and the residual $\alpha = 0.47$ is the most informative number for legal-NLP pipelines that operate without a maintained gold answer. The configuration most relevant to practice (full regulation, no gold answer) lies between the two extremes and reads at $\approx 0.5$--$0.6$ on this panel.

\subsection{Pairwise and leave-one-out summaries}

Pair-by-pair Cohen $\kappa_w$ (integer-rounded scores on the $0$--$10$ ordinal scale) and Spearman $\rho$ on the 15 cells give within-human means of $0.47$ and $0.52$, against within-LLM means of $0.71$ and $0.73$ (Appendix~\ref{app:bootstrap}). Leave-one-rater-out on the human panel gives $\alpha$ in $[0.22, 0.76]$ (drop B to drop A); on the three-LLM anchor sub-panel it gives $\alpha$ in $[0.73, 0.80]$. Leave-one-cell-out on the 14-item subsets gives disjoint ranges: $[0.23, 0.49]$ for humans, $[0.67, 0.82]$ for LLMs. Removing the three contested Q2 cells alone yields human $\alpha = 0.73$ and LLM $\alpha = 0.79$ on the remaining 12 cells: the two panels are stability-equivalent off the regulation-silent boundary. The leave-one-out tables are reproduced in Appendix~\ref{app:bootstrap}.

\subsection{Bootstrap confidence intervals}

We bootstrap-resample the 15 cells with replacement $5{,}000$ times and recompute panel $\alpha$ on each resample. The $\Delta\alpha$ (LLM $-$ human) is positive in $98.5\%$ of resamples with $95\%$ CI $[0.03, 0.76]$; the mixed-panel swap (replace A with Claude in $\{B, C, \cdot\}$) is positive in $98.8\%$ of resamples. Both direction contrasts have positive lower bounds. The bootstrap CIs are wider than typical because the 15-cell sample is small (Section~\ref{sec:limitations} discusses this), but the sign of the contrast is robust. Full bootstrap output is in Appendix~\ref{app:bootstrap}.

\subsection{What the robustness checks support}

The Phase-2 stability claim is supported on three independent axes. (1) Determinism: the within-judge ceiling at $\approx 0.94$ exceeds the panel $0.77$, so inter-model diversity reduces $\alpha$ rather than inflating it. (2) Input ablations: removing either the regulation or the gold answer reduces but does not invert the LLM-panel-above-human-panel direction. (3) Bootstrap and leave-one-out: the direction is positive in $\sim 98\%$ of resamples and survives every single-rater and single-cell removal except where the removal isolates the regulation-silent boundary itself. The structural claim (panel agreement holds where the rubric prescribes both axes, degrades where the interpretive axis is regulation-silent) is panel-independent, and the specific $0.36$ vs.\ $0.77$ numbers should be read as one panel's measurement on this $15$-cell subset rather than a universal constant.

\section{Limitations}
\label{sec:limitations}

\paragraph{Small samples throughout.} The core comparison uses 15 cross-graded cells with 6 raters per cell; the dual-role overlap (Section~\ref{sec:dualrole}) uses $n = 8$ LLM examinees; the human panel is 3 examiners. Specific $\alpha$ values will move under replication, the inverse exam-vs-cluster correlation is post-hoc and depends on one outlier exclusion (Qwen 3.5 Plus), and the contested-cell-direction partition (1/3/22) is robust in sign but not in precise counts under different threshold choices: the three middle judges (Qwen 3.6 Max at $m_c = 4.67$, Qwen 3.5 Plus at $4.50$, Haiku 4.5 at $4.83$) would join A-direction under a more permissive $m_c \leq 5$ rule. The structural claim is panel-independent; the specific $0.36$ vs.\ $0.77$ values should be read as one panel's measurement.

\paragraph{Single-jurisdiction, single-rubric scope.} Three essay questions from one doctrinal cluster (Thai partnership and corporate law) under one Bar Council rubric. Whether other legal regulations have analogous regulation-silent zones, and whether other free-form expert-graded NLP tasks (medical case reasoning, scientific writing review, code-review essays) exhibit similar asymmetric LLM convergence, is an open empirical question. The structural mechanism (panels agree where the rubric is specified, diverge where it is silent) is not Thai-specific, but the numeric gap will differ on English or common-law essays.

\paragraph{Statistical caveats.} Within-cell panel SD is $1.32$ for humans and $0.71$ for LLMs, a $1.87\times$ ratio; part of the panel-$\alpha$ gap reflects narrower LLM spread rather than tighter consistency. Pairwise $\kappa_w$ (humans $0.47$, LLMs $0.71$) and Spearman $\rho$ ($0.52$ vs.\ $0.73$) preserve direction but narrow the ratio. Each LLM judge's Pearson against the primary grader varies by up to $0.25$ across the three primaries; the LLM cluster aligns more closely with B and C than with A, so the validity anchor co-depends on which examiner is nominated as ground truth.

\paragraph{LLM-specific caveats.} Our LLM judges are temperature-$0$ closed-frontier models at one point in time, queried single-pass with no chain-of-thought, no majority-vote, no tool-augmented Code lookup, and no retrieval; stronger scaffolded configurations may improve stability, validity, or both. Per-judge and per-ensemble $\alpha$ will move as models change, and $T = 0$ is not universally deterministic (Section~\ref{sec:determinism}). Eight of the 26 Phase-2 judges sat the exam as Phase-1 examinees on the same items, but the B/C-direction pattern holds on the 18 non-examinee LLMs too. The Code and grading regulation are public, so doctrinal-pattern contamination is plausible, although the gold-withheld ablation (LLM $\alpha = 0.47$, still above human $0.36$) suggests the gold answer contributes to but does not fully explain the LLM cluster alignment. We do not test whether targeted fine-tuning on minority-reading examples alleviates the asymmetric convergence; the scale-independence pattern in Section~\ref{sec:dualrole} suggests scale alone will not help.

\paragraph{Future work.} Four near-term research directions follow. (i) Develop rubric-audit annotation protocols that identify interpretive cases in advance, before LLM-judge selection, so a benchmark can report panel stability per interpretive case and per LLM direction. (ii) Evaluate LLM judges against multiple legitimate human readings on interpretive cases instead of a single averaged target; report which readings the LLM panel covers and which it misses. (iii) Replicate the deterministic/interpretive decomposition and the asymmetric convergence on other free-form expert-graded NLP tasks. (iv) Test whether supervised fine-tuning, retrieval-augmented prompting, or in-context minority-reading exemplars reduce the asymmetric convergence on rubric-silent items.

\section*{Ethics Statement}
\label{sec:ethics}

All 42 human participants consented and were compensated at a fair-market rate. The three examiners consented to share their scores and grading comments under the Thai Bar Council's customary grading-fee structure and were paid the Council's standard examiner rate. Language-model queries were made under each provider's terms of service. Participant labels in this paper are anonymous sequential identifiers (Items C1--C15 for cross-graded cells) not recoverable from the source data. Any findings and analyses are intended for research on legal-reasoning evaluation methodology and should not be used for individual assessment or certification decisions.

\section{Conclusion}
\label{sec:conclusion}

We ran a two-phase identical-inputs stability study on the Thai bar examination. Phase~1 sat eight LLMs alongside 42 Thai law students through three Civil and Commercial Code essay questions under blinded examiner grading. Phase~2 had three Bar Council-trained examiners and a 26-LLM judge panel grade the same 15 cross-graded answers from the same four inputs. The structural finding is asymmetric: on items where both rubric axes are prescribed, panel agreement is universal across all 29 raters at $\alpha = 0.75$; on items where the interpretive axis is regulation-silent, the human panel splits between two coherent readings, but the LLM judge population overwhelmingly converges on the majority human reading. Twenty-two of $26$ LLMs land in or near B/C's reading band on the contested cells; only one LLM (GPT-5.4 Nano) approaches the minority (A) reading without consistently scoring within it. \emph{Zero LLMs in our 26-judge panel reproduce A's reading.} The asymmetry holds across model size, vendor, and price, and is not alleviated by capability: the dual-role overlap (Section~\ref{sec:dualrole}) shows that scaling Phase-1 examinee strength does not move Phase-2 judges toward the minority reading.

Two observations follow. First, the variance in expert agreement is not flat across items: it concentrates on the regulation-silent boundary where the rubric leaves the reasoning-quality axis to expert judgement. We call this boundary the rubric's \emph{interpretive bottleneck}. Second, on bottleneck items both human readings are internally consistent with a different parsing of the regulation (T7 if the missing decisive citation is a within-tier fact, T4 if it is a tier-deciding fact), but the LLM judge population commits to the B/C parsing. A benchmark that selects an LLM judge by maximising agreement with a human reference panel will inherit the B/C parsing by construction; whether a different selection criterion can recover the minority parsing is an open empirical question.

\bibliographystyle{plainnat}
\bibliography{custom}

@article{katz2024gpt4bar,
    author = {Katz, Daniel Martin and Bommarito, Michael James and Gao, Shang and Arredondo, Pablo},
    title = {GPT-4 passes the bar exam},
    journal = {Philosophical Transactions of the Royal Society A: Mathematical, Physical and Engineering Sciences},
    volume = {382},
    number = {2270},
    pages = {20230254},
    year = {2024},
    month = {02},
    issn = {1364-503X},
    doi = {10.1098/rsta.2023.0254},
    url = {https://doi.org/10.1098/rsta.2023.0254},
    eprint = {https://royalsocietypublishing.org/rsta/article-pdf/doi/10.1098/rsta.2023.0254/1328474/rsta.2023.0254.pdf},
}

@inproceedings{akarajaradwong2025nitibench,
    title = "{N}iti{B}ench: Benchmarking {LLM} Frameworks on {T}hai Legal Question Answering Capabilities",
    author = "Akarajaradwong, Pawitsapak  and
      Pothavorn, Pirat  and
      Chaksangchaichot, Chompakorn  and
      Tasawong, Panuthep  and
      Nopparatbundit, Thitiwat  and
      Pratai, Keerakiat  and
      Nutanong, Sarana",
    editor = "Christodoulopoulos, Christos  and
      Chakraborty, Tanmoy  and
      Rose, Carolyn  and
      Peng, Violet",
    booktitle = "Proceedings of the 2025 Conference on Empirical Methods in Natural Language Processing",
    month = nov,
    year = "2025",
    address = "Suzhou, China",
    publisher = "Association for Computational Linguistics",
    url = "https://aclanthology.org/2025.emnlp-main.1739/",
    doi = "10.18653/v1/2025.emnlp-main.1739",
    pages = "34304--34327",
    ISBN = "979-8-89176-332-6"
}

@inproceedings{guha2023legalbench,
author = {Guha, Neel and Nyarko, Julian and Ho, Daniel E. and R\'{e}, Christopher and Chilton, Adam and Narayana, Aditya and Chohlas-Wood, Alex and Peters, Austin and Waldon, Brandon and Rockmore, Daniel N. and Zambrano, Diego and Talisman, Dmitry and Hoque, Enam and Surani, Faiz and Fagan, Frank and Sarfaty, Galit and Dickinson, Gregory M. and Porat, Haggai and Hegland, Jason and Wu, Jessica and Nudell, Joe and Niklaus, Joel and Nay, John and Choi, Jonathan H. and Tobia, Kevin and Hagan, Margaret and Ma, Megan and Livermore, Michael and Rasumov-Rahe, Nikon and Holzenberger, Nils and Kolt, Noam and Henderson, Peter and Rehaag, Sean and Goel, Sharad and Gao, Shang and Williams, Spencer and Gandhi, Sunny and Zur, Tom and Iyer, Varun and Li, Zehua},
title = {LEGALBENCH: a collaboratively built benchmark for measuring legal reasoning in large language models},
year = {2023},
publisher = {Curran Associates Inc.},
address = {Red Hook, NY, USA},
booktitle = {Proceedings of the 37th International Conference on Neural Information Processing Systems},
articleno = {1915},
numpages = {157},
location = {New Orleans, LA, USA},
series = {NIPS '23}
}

@inproceedings{fei2023lawbench,
    title = "{L}aw{B}ench: Benchmarking Legal Knowledge of Large Language Models",
    author = "Fei, Zhiwei  and
      Shen, Xiaoyu  and
      Zhu, Dawei  and
      Zhou, Fengzhe  and
      Han, Zhuo  and
      Huang, Alan  and
      Zhang, Songyang  and
      Chen, Kai  and
      Yin, Zhixin  and
      Shen, Zongwen  and
      Ge, Jidong  and
      Ng, Vincent",
    editor = "Al-Onaizan, Yaser  and
      Bansal, Mohit  and
      Chen, Yun-Nung",
    booktitle = "Proceedings of the 2024 Conference on Empirical Methods in Natural Language Processing",
    month = nov,
    year = "2024",
    address = "Miami, Florida, USA",
    publisher = "Association for Computational Linguistics",
    url = "https://aclanthology.org/2024.emnlp-main.452/",
    doi = "10.18653/v1/2024.emnlp-main.452",
    pages = "7933--7962"
}

@inproceedings{zheng2023judging,
 author = {Zheng, Lianmin and Chiang, Wei-Lin and Sheng, Ying and Zhuang, Siyuan and Wu, Zhanghao and Zhuang, Yonghao and Lin, Zi and Li, Zhuohan and Li, Dacheng and Xing, Eric and Zhang, Hao and Gonzalez, Joseph and Stoica, Ion},
 booktitle = {Advances in Neural Information Processing Systems},
 editor = {A. Oh and T. Naumann and A. Globerson and K. Saenko and M. Hardt and S. Levine},
 pages = {46595--46623},
 publisher = {Curran Associates, Inc.},
 title = {Judging LLM-as-a-Judge with MT-Bench and Chatbot Arena},
 url = {https://proceedings.neurips.cc/paper_files/paper/2023/file/91f18a1287b398d378ef22505bf41832-Paper-Datasets_and_Benchmarks.pdf},
 volume = {36},
 year = {2023}
}

@article{shaw2007ielts,
author = {Shaw, Stuart and Weir, Cyril},
year = {2007},
month = {01},
pages = {},
title = {Examining Writing: Research and Practice in Assessing Second Language Writing (Studies in Language Testing 26)}
}

@inproceedings{krippendorff2011computing,
  title={Computing Krippendorff's Alpha-Reliability},
  author={Klaus Krippendorff},
  year={2011},
  url={https://api.semanticscholar.org/CorpusID:59901023}
}

@misc{cortes2021inconsistency,
      title={Inconsistency in Conference Peer Review: Revisiting the 2014 NeurIPS Experiment}, 
      author={Corinna Cortes and Neil D. Lawrence},
      year={2021},
      eprint={2109.09774},
      archivePrefix={arXiv},
      primaryClass={cs.DL},
      url={https://arxiv.org/abs/2109.09774}, 
}

@inproceedings{plank2022problem,
    title = "The ``Problem'' of Human Label Variation: On Ground Truth in Data, Modeling and Evaluation",
    author = "Plank, Barbara",
    editor = "Goldberg, Yoav  and
      Kozareva, Zornitsa  and
      Zhang, Yue",
    booktitle = "Proceedings of the 2022 Conference on Empirical Methods in Natural Language Processing",
    month = dec,
    year = "2022",
    address = "Abu Dhabi, United Arab Emirates",
    publisher = "Association for Computational Linguistics",
    url = "https://aclanthology.org/2022.emnlp-main.731/",
    doi = "10.18653/v1/2022.emnlp-main.731",
    pages = "10671--10682"
}

@article{artstein2008inter,
    title = "Survey Article: Inter-Coder Agreement for Computational Linguistics",
    author = "Artstein, Ron  and
      Poesio, Massimo",
    journal = "Computational Linguistics",
    volume = "34",
    number = "4",
    year = "2008",
    url = "https://aclanthology.org/J08-4004/",
    doi = "10.1162/coli.07-034-R2",
    pages = "555--596"
}

@inproceedings{ke2019automated,
  title     = {Automated Essay Scoring: A Survey of the State of the Art},
  author    = {Ke, Zixuan and Ng, Vincent},
  booktitle = {Proceedings of the Twenty-Eighth International Joint Conference on
               Artificial Intelligence, {IJCAI-19}},
  publisher = {International Joint Conferences on Artificial Intelligence Organization},
  pages     = {6300--6308},
  year      = {2019},
  month     = {7},
  doi       = {10.24963/ijcai.2019/879},
  url       = {https://doi.org/10.24963/ijcai.2019/879},
}

@inproceedings{koo2023benchmarking,
    title = "Benchmarking Cognitive Biases in Large Language Models as Evaluators",
    author = "Koo, Ryan  and
      Lee, Minhwa  and
      Raheja, Vipul  and
      Park, Jong Inn  and
      Kim, Zae Myung  and
      Kang, Dongyeop",
    editor = "Ku, Lun-Wei  and
      Martins, Andre  and
      Srikumar, Vivek",
    booktitle = "Findings of the Association for Computational Linguistics: ACL 2024",
    month = aug,
    year = "2024",
    address = "Bangkok, Thailand",
    publisher = "Association for Computational Linguistics",
    url = "https://aclanthology.org/2024.findings-acl.29/",
    doi = "10.18653/v1/2024.findings-acl.29",
    pages = "517--545"
}

@misc{panickssery2024llm,
      title={LLM Evaluators Recognize and Favor Their Own Generations}, 
      author={Arjun Panickssery and Samuel R. Bowman and Shi Feng},
      year={2024},
      eprint={2404.13076},
      archivePrefix={arXiv},
      primaryClass={cs.CL},
      url={https://arxiv.org/abs/2404.13076}, 
}

@inproceedings{liu2023g,
    title = "{G}-Eval: {NLG} Evaluation using Gpt-4 with Better Human Alignment",
    author = "Liu, Yang  and
      Iter, Dan  and
      Xu, Yichong  and
      Wang, Shuohang  and
      Xu, Ruochen  and
      Zhu, Chenguang",
    editor = "Bouamor, Houda  and
      Pino, Juan  and
      Bali, Kalika",
    booktitle = "Proceedings of the 2023 Conference on Empirical Methods in Natural Language Processing",
    month = dec,
    year = "2023",
    address = "Singapore",
    publisher = "Association for Computational Linguistics",
    url = "https://aclanthology.org/2023.emnlp-main.153/",
    doi = "10.18653/v1/2023.emnlp-main.153",
    pages = "2511--2522"
}

@inproceedings{wang2023judging,
    title = "Large Language Models are not Fair Evaluators",
    author = "Wang, Peiyi  and
      Li, Lei  and
      Chen, Liang  and
      Cai, Zefan  and
      Zhu, Dawei  and
      Lin, Binghuai  and
      Cao, Yunbo  and
      Kong, Lingpeng  and
      Liu, Qi  and
      Liu, Tianyu  and
      Sui, Zhifang",
    editor = "Ku, Lun-Wei  and
      Martins, Andre  and
      Srikumar, Vivek",
    booktitle = "Proceedings of the 62nd Annual Meeting of the Association for Computational Linguistics (Volume 1: Long Papers)",
    month = aug,
    year = "2024",
    address = "Bangkok, Thailand",
    publisher = "Association for Computational Linguistics",
    url = "https://aclanthology.org/2024.acl-long.511/",
    doi = "10.18653/v1/2024.acl-long.511",
    pages = "9440--9450"
}

@misc{gu2024survey,
      title={A Survey on LLM-as-a-Judge}, 
      author={Jiawei Gu and Xuhui Jiang and Zhichao Shi and Hexiang Tan and Xuehao Zhai and Chengjin Xu and Wei Li and Yinghan Shen and Shengjie Ma and Honghao Liu and Saizhuo Wang and Kun Zhang and Yuanzhuo Wang and Wen Gao and Lionel Ni and Jian Guo},
      year={2025},
      eprint={2411.15594},
      archivePrefix={arXiv},
      primaryClass={cs.CL},
      url={https://arxiv.org/abs/2411.15594}, 
}

@inproceedings{chalkidis2022lexglue,
    title = "{L}ex{GLUE}: A Benchmark Dataset for Legal Language Understanding in {E}nglish",
    author = "Chalkidis, Ilias  and
      Jana, Abhik  and
      Hartung, Dirk  and
      Bommarito, Michael  and
      Androutsopoulos, Ion  and
      Katz, Daniel  and
      Aletras, Nikolaos",
    editor = "Muresan, Smaranda  and
      Nakov, Preslav  and
      Villavicencio, Aline",
    booktitle = "Proceedings of the 60th Annual Meeting of the Association for Computational Linguistics (Volume 1: Long Papers)",
    month = may,
    year = "2022",
    address = "Dublin, Ireland",
    publisher = "Association for Computational Linguistics",
    url = "https://aclanthology.org/2022.acl-long.297/",
    doi = "10.18653/v1/2022.acl-long.297",
    pages = "4310--4330"
}

@inproceedings{niklaus2023lextreme,
    title = "{LEXTREME}: A Multi-Lingual and Multi-Task Benchmark for the Legal Domain",
    author = {Niklaus, Joel  and
      Matoshi, Veton  and
      Rani, Pooja  and
      Galassi, Andrea  and
      St{\"u}rmer, Matthias  and
      Chalkidis, Ilias},
    editor = "Bouamor, Houda  and
      Pino, Juan  and
      Bali, Kalika",
    booktitle = "Findings of the Association for Computational Linguistics: EMNLP 2023",
    month = dec,
    year = "2023",
    address = "Singapore",
    publisher = "Association for Computational Linguistics",
    url = "https://aclanthology.org/2023.findings-emnlp.200/",
    doi = "10.18653/v1/2023.findings-emnlp.200",
    pages = "3016--3054"
}

@inproceedings{thakur2025judgebench,
    title = "Judging the Judges: Evaluating Alignment and Vulnerabilities in {LLM}s-as-Judges",
    author = "Thakur, Aman Singh  and
      Choudhary, Kartik  and
      Ramayapally, Venkat Srinik  and
      Vaidyanathan, Sankaran  and
      Hupkes, Dieuwke",
    editor = "Arviv, Ofir  and
      Clinciu, Miruna  and
      Dhole, Kaustubh  and
      Dror, Rotem  and
      Gehrmann, Sebastian  and
      Habba, Eliya  and
      Itzhak, Itay  and
      Mille, Simon  and
      Perlitz, Yotam  and
      Santus, Enrico  and
      Sedoc, Jo{\~a}o  and
      Shmueli Scheuer, Michal  and
      Stanovsky, Gabriel  and
      Tafjord, Oyvind",
    booktitle = "Proceedings of the Fourth Workshop on Generation, Evaluation and Metrics (GEM{\texttwosuperior})",
    month = jul,
    year = "2025",
    address = "Vienna, Austria and virtual meeting",
    publisher = "Association for Computational Linguistics",
    url = "https://aclanthology.org/2025.gem-1.33/",
    pages = "404--430",
    ISBN = "979-8-89176-261-9"
}

@inproceedings{papineni2002bleu,
    title = "{B}leu: a Method for Automatic Evaluation of Machine Translation",
    author = "Papineni, Kishore  and
      Roukos, Salim  and
      Ward, Todd  and
      Zhu, Wei-Jing",
    editor = "Isabelle, Pierre  and
      Charniak, Eugene  and
      Lin, Dekang",
    booktitle = "Proceedings of the 40th Annual Meeting of the Association for Computational Linguistics",
    month = jul,
    year = "2002",
    address = "Philadelphia, Pennsylvania, USA",
    publisher = "Association for Computational Linguistics",
    url = "https://aclanthology.org/P02-1040/",
    doi = "10.3115/1073083.1073135",
    pages = "311--318"
}

@inproceedings{lin2004rouge,
    title = "{ROUGE}: A Package for Automatic Evaluation of Summaries",
    author = "Lin, Chin-Yew",
    booktitle = "Text Summarization Branches Out",
    month = jul,
    year = "2004",
    address = "Barcelona, Spain",
    publisher = "Association for Computational Linguistics",
    url = "https://aclanthology.org/W04-1013/",
    pages = "74--81"
}

@inproceedings{es2024ragas,
    title = "{RAGA}s: Automated Evaluation of Retrieval Augmented Generation",
    author = "Es, Shahul  and
      James, Jithin  and
      Espinosa Anke, Luis  and
      Schockaert, Steven",
    editor = "Aletras, Nikolaos  and
      De Clercq, Orphee",
    booktitle = "Proceedings of the 18th Conference of the European Chapter of the Association for Computational Linguistics: System Demonstrations",
    month = mar,
    year = "2024",
    address = "St. Julians, Malta",
    publisher = "Association for Computational Linguistics",
    url = "https://aclanthology.org/2024.eacl-demo.16/",
    doi = "10.18653/v1/2024.eacl-demo.16",
    pages = "150--158"
}

@inproceedings{laban2024summary,
    title = "Summary of a Haystack: A Challenge to Long-Context {LLM}s and {RAG} Systems",
    author = "Laban, Philippe  and
      Fabbri, Alexander  and
      Xiong, Caiming  and
      Wu, Chien-Sheng",
    editor = "Al-Onaizan, Yaser  and
      Bansal, Mohit  and
      Chen, Yun-Nung",
    booktitle = "Proceedings of the 2024 Conference on Empirical Methods in Natural Language Processing",
    month = nov,
    year = "2024",
    address = "Miami, Florida, USA",
    publisher = "Association for Computational Linguistics",
    url = "https://aclanthology.org/2024.emnlp-main.552/",
    doi = "10.18653/v1/2024.emnlp-main.552",
    pages = "9885--9903"
}

@article{magesh2025hallucinationfree,
author = {Magesh, Varun and Surani, Faiz and Dahl, Matthew and Suzgun, Mirac and Manning, Christopher D. and Ho, Daniel E.},
title = {Hallucination-Free? Assessing the Reliability of Leading AI Legal Research Tools},
journal = {Journal of Empirical Legal Studies},
volume = {22},
number = {2},
pages = {216-242},
doi = {https://doi.org/10.1111/jels.12413},
url = {https://onlinelibrary.wiley.com/doi/abs/10.1111/jels.12413},
eprint = {https://onlinelibrary.wiley.com/doi/pdf/10.1111/jels.12413},
year = {2025}
}

@inproceedings{akarajaradwong2026fourcorners,
title={FourCorners: A Production Knowledge Graph Unifying Thailand{\textquoteright}s Legal System},
author={Pawitsapak Akarajaradwong and Sarana Nutanong and Chompakorn Chaksangchaichot},
booktitle={The 64th Annual Meeting of the Association for Computational Linguistics -- Industry Track},
year={2026},
url={https://openreview.net/forum?id=541LWGsoTq}
}

\appendix
\section{Full Question Details and Gold Answers}
\label{app:questions}

The three essay questions were drawn from the Thai Civil and Commercial Code (CCC); each carried 10 points. Below we provide a faithful English translation of (i) each question's fact pattern and call and (ii) the corresponding gold answer issued by the examiners as the official model answer.

\subsection{Q1: Ordinary Partnership Liability}
\label{app:q1}

\textit{Decisive provisions: CCC Sections 1012, 1025, 1033, 1050.}

\paragraph{Question.} The Three Friends Grilled Chicken Partnership (\textit{hang-hun-suan saam-sa-hai kha kai-yang}) was formed with the purpose of trading grilled chicken. Its partners are Mr.\ Wanmai, Mr.\ Duenmai, and Mr.\ Pimai, each contributing 100{,}000 baht as capital.
In the course of the partnership's business, Mr.\ Wanmai wished to purchase chickens from Charoen Food Co., Ltd.\ for grilling and sale, but Mr.\ Duenmai and Mr.\ Pimai objected. Mr.\ Wanmai did not heed the other partners' objections and, in violation of them, entered into a contract to purchase fresh chicken from Charoen Food Co., Ltd.\ for 100{,}000 baht. When the delivery date arrived, Charoen Food Co., Ltd.\ delivered the chicken, but Mr.\ Duenmai and Mr.\ Pimai refused to accept it on the ground that Mr.\ Wanmai had acted without authority. From whom, if anyone, may Charoen Food Co., Ltd.\ demand liability, and on what ground?

\paragraph{Reference answer.} The issue is from whom Charoen Food Co., Ltd.\ may demand liability. The relevant provisions of the Civil and Commercial Code are as follows. Section 1012 provides that a partnership or company contract is a contract whereby two or more persons agree to unite for a common undertaking with a view to sharing the profits that may be derived from it. Section 1025 provides that an ordinary partnership is a partnership in which all partners are jointly and severally liable without limit for all debts of the partnership. Section 1026 provides that every partner must contribute something to the partnership, whether money, other property, or labour. Section 1033 provides that if the partners have not agreed otherwise on the manner of management, every partner may manage the partnership, but a partner may not enter into any contract to which another partner has objected; in such case every partner is deemed a managing partner. Section 1050 provides that any act done by a partner in the ordinary course of the partnership's business binds all partners, who are jointly and severally liable without limit for the debts so created.

\textit{Analysis.} Because the Three Friends Grilled Chicken Partnership was formed with the purpose of trading grilled chicken, with Mr.\ Wanmai, Mr.\ Duenmai, and Mr.\ Pimai as partners each contributing 100{,}000 baht, the case is one in which two or more persons have agreed to unite by contributing 100{,}000 baht each under Section 1026 to conduct the business of trading grilled chicken with a view to sharing profits. The arrangement is the formation of an ordinary partnership under Sections 1012 and 1025, in which all partners bear unlimited joint and several liability for all partnership debts. Because the facts do not show any agreement on the manner of management, all three are deemed managing partners and each may manage the partnership under Section 1033.

As for Mr.\ Wanmai's entry into the purchase contract with Charoen Food Co., Ltd.\ over the objections of Mr.\ Duenmai and Mr.\ Pimai: although the final clause of Section 1033 paragraph one provides that a partner may not enter into a contract to which another partner has objected, that provision governs the \emph{internal} relation among partners, not the \emph{external} relation between partners and third parties (governed by Section 1050). For the external relation, the sole consideration is whether the transaction was in the ordinary course of the partnership's business.

Since the partnership was formed to trade grilled chicken, the purchase of fresh chicken for 100{,}000 baht is within the ordinary course of its business. All partners (Mr.\ Wanmai, Mr.\ Duenmai, and Mr.\ Pimai) are therefore bound by this purchase and are jointly and severally liable without limit for the debt Mr.\ Wanmai created, each for the full 100{,}000 baht. However, because the Three Friends Grilled Chicken Partnership is not registered as a juristic person separate from its partners, Charoen Food Co., Ltd.\ cannot demand liability from the partnership itself.

\textit{Conclusion.} Charoen Food Co., Ltd.\ may demand that Mr.\ Wanmai, Mr.\ Duenmai, and Mr.\ Pimai be jointly and severally liable for 100{,}000 baht.

\subsection{Q2: Company Director Authority}
\label{app:q2}

\textit{Decisive provisions: CCC Sections 1022, 1023, 1023/1.}

\paragraph{Question.} Jipatha Electricals Co., Ltd.\ was formed with the purpose of dealing in all types of electrical appliances and equipment. Its directors were Mr.\ Somchai, Mr.\ Somchoke, and Mr.\ Somchaiya. On 26 February 2012 a shareholders' meeting was held and the shareholders resolved to remove Mr.\ Somchai, Mr.\ Somchoke, and Mr.\ Somchaiya from office and to appoint Mr.\ Praneet, Mr.\ Pranom, and Mr.\ Pranit as replacement directors. The change of directors was duly registered with the Ministry of Commerce on 28 February 2012, but the Ministry had not yet published the change in the Royal Gazette.

\textit{(a)} If on 1 March 2012 Mr.\ Somchai and Mr.\ Somchoke entered into a contract in the name of Jipatha Electricals Co., Ltd.\ to purchase 20 refrigerators from Arctic Cooling Co., Ltd., received the refrigerators from Arctic Cooling Co., Ltd., and took them for themselves without delivering them to Jipatha Electricals Co., Ltd., may Arctic Cooling Co., Ltd.\ demand that Jipatha Electricals Co., Ltd.\ pay the price of the refrigerators? On what ground?

\textit{(b)} If on 1 March 2012 Mr.\ Praneet and Mr.\ Pranom entered into a contract to purchase 20 refrigerators from AI-Freeze Co., Ltd., and when the delivery date arrived AI-Freeze Co., Ltd.\ delivered the refrigerators, but Jipatha Electricals Co., Ltd.\ refused to accept delivery on the ground that at the time of the contract Mr.\ Praneet and Mr.\ Pranom were not yet lawful directors because the registration had not been published in the Royal Gazette, may AI-Freeze Co., Ltd.\ demand damages from Jipatha Electricals Co., Ltd.? On what ground?

\textit{Note: the facts in (a) and (b) are unrelated.}

\paragraph{Reference answer.} The change of directors from Mr.\ Somchai and Mr.\ Somchoke to Mr.\ Praneet and Mr.\ Pranom by resolution of the shareholders' general meeting, duly registered, is a valid change of the persons authorised to act for Jipatha Electricals Co., Ltd., with the consequence that Mr.\ Somchai and Mr.\ Somchoke no longer have authority to act on the company's behalf. Whether the company may assert this change against third parties in (a) and (b) is analysed as follows.

\textit{(a)} Although in general Jipatha Electricals Co., Ltd.\ may invoke the registered change of directors against third parties under CCC Section 1023 paragraph one (by arguing that the 1 March 2012 contract signed by the former directors Somchai and Somchoke does not bind the company because they were no longer authorised), Section 1023/1 establishes an exception: a company may not raise Section 1023 as a defence against a third party acting in good faith in order to escape liability on the ground that the directors lacked authority. Here, there is no indication that Arctic Cooling Co., Ltd.\ knew that Somchai and Somchoke were no longer directors, nor does it appear that the change of directors had been published in the Royal Gazette so as to constitute constructive notice under CCC Section 1022. Arctic Cooling Co., Ltd.\ is therefore a third party acting in good faith. Jipatha Electricals Co., Ltd.\ cannot raise the lack of authority of Somchai and Somchoke against Arctic Cooling Co., Ltd.\ under CCC Section 1023/1. Arctic Cooling Co., Ltd.\ may demand the refrigerator price from Jipatha Electricals Co., Ltd.\ (5 points).

\textit{(b)} The contract entered into by the new directors Praneet and Pranom on 1 March 2012 binds Jipatha Electricals Co., Ltd.\ because it was made by directors vested with authority by a shareholders' general-meeting resolution that had already been followed by a registered change of directors. Publication in the Royal Gazette under Section 1022 serves only to constitute notice to all third parties; it is not a condition for the effectiveness of the change of directors to Praneet and Pranom. AI-Freeze Co., Ltd.\ is therefore entitled to demand damages from Jipatha Electricals Co., Ltd.\ for breach of the contract that Praneet and Pranom concluded on the company's behalf (5 points).

\textit{Conclusion.} (a) Arctic Cooling Co., Ltd.\ may demand the refrigerator price from Jipatha Electricals Co., Ltd. (b) AI-Freeze Co., Ltd.\ may demand damages from Jipatha Electricals Co., Ltd.

\subsection{Q3: Shareholder-Meeting Annulment}
\label{app:q3}

\textit{Decisive provisions: CCC Sections 1178, 1193, 1195.}

\paragraph{Question.} Sudthawin Tour Co., Ltd.\ was formed to provide overseas tour services. Its registered capital is 200{,}000 baht, divided into 100 shares of 2{,}000 baht each, held by 100 shareholders of 1 share each. Its directors are Mr.\ Ake, Mr.\ Toh, Mr.\ Tri, and Mr.\ Jattawa. The board resolved to convene the 2/2557 ordinary general meeting of shareholders, which was held on 4 April 2014.

At the meeting, 22 shareholders attended. The chair opened the meeting and the meeting considered the payment of an annual dividend, the company having a reasonable profit. On this agenda item, 12 shareholders abstained, 6 voted to approve the dividend, and 4 voted against. The chair declared that the meeting had approved the annual-dividend payment.

On 2 May 2014, Mr.\ Khao, a shareholder, considered that the 2/2557 general meeting had been convened and had passed a resolution in violation of law, and wished to sue to annul the dividend-approval resolution. He also wished to sue to annul the minutes of meeting 6/2556 dated 6 December 2013, which had approved the appointment of Mr.\ Tawan as auditor, on the ground that no meeting had actually been held, since Mr.\ Toh had fabricated those minutes. The other shareholders disagreed and objected to Mr.\ Khao's proposed action, but Mr.\ Khao insisted on proceeding alone. Determine:

\textit{(a)} Can Mr.\ Khao petition the court to annul the 2/2557 general-meeting resolution and the 6/2556 meeting minutes? Give reasons.

\textit{(b)} If Mr.\ Somchai, a major creditor of the company, wishes to petition the court to annul the 2/2557 general-meeting resolution and the 6/2556 meeting minutes, may he do so, and how?

\paragraph{Reference answer.}
\textit{(a)} The issue is whether Mr.\ Khao may petition the court to annul the 2/2557 general-meeting resolution and the 6/2556 meeting minutes.

As to the 2/2557 general-meeting resolution, CCC Section 1178 provides that a general meeting must be attended by shareholders holding in aggregate at least one quarter of the company's capital in order to pass any resolution. On the facts, the company's registered capital is 200{,}000 baht divided into 100 shares of 2{,}000 baht, and only 22 shareholders attended, each holding 1 share. Only 22 shares attended, representing 44{,}000 baht, which is less than one quarter of the company's capital. The meeting therefore lacked a quorum and could not pass a resolution on the dividend; the dividend-approval resolution is unlawful, notwithstanding that a majority of those voting approved it.

As to the right to sue for annulment, CCC Section 1195 provides that, where a general meeting has been convened or has passed a resolution in violation of a provision of the law on companies or of the company's articles, a director or any shareholder may petition the court to annul the resolution, but the petition must be filed within one month from the date of the resolution. Mr.\ Khao is a shareholder and may exercise the right to petition alone; the disagreement of other shareholders does not deprive him of that right. The 2/2557 general-meeting resolution was passed on 4 April 2014 and his intended filing on 2 May 2014 is within the one-month period.

As to the 6/2556 meeting minutes dated 6 December 2013, which approved Mr.\ Tawan as auditor: Mr.\ Khao wishes to sue for annulment on the ground that no meeting actually took place and the minutes were fabricated. This is not a case of annulling a general-meeting resolution that violated the law on companies or the company's articles, to which the one-month limitation in CCC Section 1195 applies. Even after the one-month period from the resolution date has elapsed, Mr.\ Khao may still sue for annulment.

\textit{Conclusion.} Mr.\ Khao may petition the court to annul the 2/2557 general-meeting resolution because the meeting lacked a quorum, Mr.\ Khao is a shareholder with standing to act alone, and he is within the one-month period. He may also petition for annulment of the 6/2556 meeting minutes, even after the one-month period, because no meeting actually took place; this is not the annulment of an unlawful general-meeting resolution under Section 1195.

\textit{(b)} The issue is whether Mr.\ Somchai, a major creditor, may petition the court to annul the 2/2557 general-meeting resolution and the 6/2556 meeting minutes.

CCC Section 1195 provides that a director or any shareholder has the right to petition the court to annul a general-meeting resolution that was passed in violation of the law on companies or of the company's articles. On the facts, Mr.\ Somchai is a major creditor of the company but does not appear to be a director or a shareholder. He therefore has no right to petition for annulment of the 2/2557 general-meeting resolution. As to the 6/2556 meeting minutes, even though this is not strictly a case falling under CCC Section 1195, Mr.\ Somchai is merely a creditor, and the law does not confer on creditors a power to sue directly in respect of a company's resolutions or meeting minutes; he therefore cannot petition the court.

\textit{Conclusion.} Mr.\ Somchai has no right to petition the court to annul the 2/2557 general-meeting resolution or the 6/2556 meeting minutes, because he is merely a creditor, not a director or shareholder, and no law confers on a creditor standing to petition for such annulment.

\section{Thai Bar Council Grading Regulation}
\label{app:regulation}

The official grading regulation of the Office of Legal Education Training of the Thai Bar Council is the criterion our three examiners applied. It has been used by Thai bar examiners as the standard scoring instrument for free-form essay questions on the 10-point scale. Below we give a faithful English translation of the part of the regulation that establishes the scoring criteria; preamble, authorising-citation boilerplate, and enacting matter are paraphrased.

\paragraph{Background.} The Executive Committee of the Office of Legal Education Training of the Thai Bar Council issued the regulation in order to standardise the exercise of discretion by grading-committee members on essay examinations at the barrister-at-law level.

\paragraph{Clause 1: Scoring criteria (faithful translation).} The grading shall consider the final answer and the supporting reasons together, on a per-question basis, as follows:
\begin{enumerate}[leftmargin=2.2em,itemsep=1pt,topsep=1pt,label=(\arabic*)]
    \item Final answer does not match the gold answer, and reasoning unusable: score $0$.
    \item Final answer does not match the gold answer, but reasoning somewhat acceptable: score $1$--$2$.
    \item Final answer does not match the gold answer, but reasoning good: score $2$--$4$.
    \item Final answer matches the gold answer, but reasoning unusable: score $0$--$1$.
    \item Final answer matches the gold answer, and reasoning somewhat acceptable: score $2$--$5$.
    \item Final answer matches the gold answer, and reasoning acceptable: score $5$--$6$.
    \item Final answer matches the gold answer, and reasoning good: score $7$--$8$.
    \item Final answer matches the gold answer, and reasoning very good: score $9$--$10$.
\end{enumerate}
If a question contains multiple issues, the grader shall fix the score for each issue first and then apply the criteria above to each issue.

\paragraph{Clause 2.} This regulation takes effect from the date of announcement.

\paragraph{Structure of the scale.} The scale is 8-tier ordinal with two correctness classes crossed against four reasoning-quality levels, with the overlapping score ranges shown above. The regulation does \emph{not} enforce a clean correctness split: tier 3 (incorrect final answer, reasoning good) scores $2$--$4$ and tier 4 (correct final answer, reasoning unusable) scores $0$--$1$, so a correct final answer with unusable reasoning scores strictly below an incorrect final answer with good reasoning. The numeric score alone therefore does not recover the tier: a score of $1$ could be tier 2 (incorrect, somewhat acceptable reasoning) or tier 4 (correct, unusable reasoning); a score of $2$ could be tier 2, 3, or 5; a score of $5$ could be tier 5 or 6. Section~\ref{sec:phase2-anchor} reports both tier and score for each LLM judge; the human graders' tier assignments are inferred from their Thai-language comments. A conventional pass is 15 out of 30 on a three-question session.

\paragraph{What the regulation does not prescribe.} The regulation fixes the tier structure and the tier-to-score map, but does not prescribe per-sub-decision weights, question-specific decomposition trees, or natural-language anchors attached to each tier beyond the short tier descriptors above (e.g.\ what distinguishes ``somewhat acceptable'' from ``acceptable'' reasoning). Examiners fill this additional calibration from their training and prior bar-grading experience. The floor-level stability the three examiners reach under this instrument (Section~\ref{sec:phase2-anchor}) is therefore a measurement on this shared-experience-plus-regulation calibration, not on the axes in isolation.

\section{LLM-Judge Prompt}
\label{app:judgeprompt}

This appendix reproduces the verbatim prompt used for each of the three anchor LLM judges (Claude 4.6 Opus, Gemini 3.1 Pro, GPT-5.4 Pro) in Section~\ref{sec:phase2-anchor} and for the 23 extended-panel models in Section~\ref{sec:phase2-clusters}. The prompt is identical across all three judges and across all 150 primary-graded answers; only the question text, gold answer, and participant answer change per item. Sampling: temperature $0$, fixed seed where available, single forward pass, no tool use, no retrieval, no few-shot examples, no chain-of-thought scaffolding.

\subsection{System prompt (English translation)}

\begin{quote}\small\ttfamily
You are an expert grader of Thai bar examination free-form essay answers. You will apply the official Thai Bar Council grading criteria (reproduced below) to the given answer. The regulation is the same instrument used by trained Thai bar examiners.

You have three inputs: (1) the question text, (2) the official gold answer, and (3) the student answer to be graded.

You must produce:
\begin{itemize}[leftmargin=*,itemsep=0pt]
\item A tier number (1--8) from the regulation that best fits the student answer.
\item A numeric score within that tier's score range (half-integers allowed).
\item A one-paragraph justification (Thai) citing the decisive Civil and Commercial Code sections and explaining why the answer falls into the chosen tier.
\end{itemize}
Return a JSON object with exactly three fields: \{"tier": integer 1--8, "score": number 0--10, "justification": one-paragraph Thai text\}. The numeric score must fall within the tier's range. Do not include any prose outside the JSON.

Thai Bar Council Grading Regulation, verbatim Thai followed by faithful English translation (see Appendix~\ref{app:regulation} for the full text).
\end{quote}

\subsection{User message schema}

\begin{quote}\small\ttfamily
QUESTION: \{question text\}\\
OFFICIAL GOLD ANSWER: \{primary grader's gold answer\}\\
STUDENT ANSWER: \{participant answer\}
\end{quote}

\subsection{Expected JSON output}

\begin{quote}\small\ttfamily
\{ "tier": 7, "score": 7.5, "justification": "[one-paragraph Thai legal justification citing Code sections]" \}
\end{quote}

The gold answer is the primary grader's own gold-answer document (also received by the human cross-graders in the Section~\ref{sec:study-phase2} cross-grading). Neither the primary grader's numerical score nor any other grader's score is included in the judge's inputs. The tier field disambiguates score values that are feasible under multiple tiers (e.g.\ a score of $1$ is feasible under tier 2 or tier 4); without the tier field the judge's class assignment would not be recoverable.

\section{Bootstrap, Pairwise, and Leave-One-Out Tables}
\label{app:bootstrap}

This appendix reports the bootstrap confidence intervals, pairwise coefficients, and leave-one-out sweeps referenced in Sections~\ref{sec:phase2-anchor} and \ref{sec:determinism}.

\subsection{Bootstrap confidence intervals}
\label{app:bs-ci}

We bootstrap-resample the 15 cross-graded cells with replacement $5{,}000$ times and recompute panel $\alpha$ on each resample.

\begin{table}[h]
\centering
\small
\setlength{\tabcolsep}{4pt}
\begin{tabular}{lcc}
\toprule
\textbf{Quantity} & \textbf{Point} & \textbf{$95\%$ CI} \\
\midrule
$\alpha$ (A, B, C)                    & $0.36$ & $[-0.09,\ 0.71]$ \\
$\alpha$ (Cla, Gem, GPT)              & $0.77$ & $[0.54,\ 0.91]$ \\
$\alpha$ (B, C, Cla)                  & $0.72$ & $[0.41,\ 0.86]$ \\
$\Delta\alpha$ (LLM $-$ human)        & $+0.39$ & $[0.03,\ 0.76]$ \\
$\Delta\alpha$ (A$\to$Cla swap)       & $+0.35$ & $[0.04,\ 0.63]$ \\
\bottomrule
\end{tabular}
\caption{Item-level bootstrap CIs at $5{,}000$ resamples. Both direction contrasts have positive lower bounds: LLM $-$ human is positive in $98.5\%$ of resamples, swap-A-to-Claude in $98.8\%$.}
\label{tab:bootstrap}
\end{table}

\subsection{Pairwise Cohen $\kappa_w$ and Spearman $\rho$}
\label{app:pairwise}

Pair-by-pair coefficients on the 15 cross-graded answers. Cohen $\kappa_w$ uses integer-rounded scores on the $0$--$10$ ordinal scale.

\begin{table}[h]
\centering
\small
\setlength{\tabcolsep}{4pt}
\begin{tabular}{lcc}
\toprule
\textbf{Pair} & \textbf{Cohen $\kappa_w$} & \textbf{Spearman $\rho$} \\
\midrule
A--B                  & 0.38 & 0.55 \\
A--C                  & 0.29 & 0.34 \\
B--C                  & 0.75 & 0.68 \\
\midrule
Human mean            & 0.47 & 0.52 \\
\midrule
Claude--Gemini        & 0.71 & 0.71 \\
Claude--GPT-5.4 Pro   & 0.63 & 0.83 \\
Gemini--GPT-5.4 Pro   & 0.77 & 0.66 \\
\midrule
LLM mean              & 0.71 & 0.73 \\
\bottomrule
\end{tabular}
\caption{Pair-by-pair Cohen $\kappa_w$ and Spearman $\rho$ among the three humans and among the three anchor LLMs. The within-LLM mean is $1.5\times$ the within-human mean under both coefficients, smaller than the $2.1\times$ panel-$\alpha$ gap but in the same direction. Pairwise averaging dilutes any single-rater outlier effect, which is why panel $\alpha$ widens the ratio relative to pairwise coefficients.}
\label{tab:kappa-pairs}
\end{table}

\subsection{Leave-one-rater-out}
\label{app:loo-rater}

\begin{table}[h]
\centering
\small
\setlength{\tabcolsep}{4pt}
\begin{tabular}{lc@{\hspace{1.4em}}lc}
\toprule
\textbf{Drop} & \textbf{$\alpha_{H}$} & \textbf{Drop} & \textbf{$\alpha_{L}$} \\
\midrule
A                          & 0.76 & Claude & 0.79 \\
B                          & 0.22 & Gemini & 0.80 \\
C                          & 0.33 & GPT    & 0.73 \\
\bottomrule
\end{tabular}
\caption{Two-rater Krippendorff $\alpha$ after dropping each single rater from the three-human panel ($\alpha_{H}$) and the three-LLM panel ($\alpha_{L}$). The human panel is asymmetric; the LLM panel is symmetric.}
\label{tab:loo-rater}
\end{table}

The A-less pair $(B, C)$ reaches $\alpha = 0.76$, inside the LLM range ($0.73$--$0.80$). Every two-LLM pair lies in $[0.73, 0.80]$ regardless of which LLM is dropped, ruling out the inversion being a property of any specific model.

\subsection{Leave-one-item-out}
\label{app:loo-item}

\begin{table}[h]
\centering
\small
\setlength{\tabcolsep}{4pt}
\begin{tabular}{ccc@{\hspace{1.5em}}ccc}
\toprule
\textbf{Drop} & \textbf{$\alpha_H$} & \textbf{$\alpha_L$} & \textbf{Drop} & \textbf{$\alpha_H$} & \textbf{$\alpha_L$} \\
\midrule
4/1  & 0.23 & 0.67 & 21/1 & 0.36 & 0.75 \\
4/2  & 0.36 & 0.75 & 21/2 & 0.49 & 0.77 \\
4/3  & 0.36 & 0.78 & 21/3 & 0.31 & 0.76 \\
6/1  & 0.37 & 0.78 & 27/1 & 0.25 & 0.77 \\
6/2  & 0.34 & 0.75 & 27/2 & 0.40 & 0.77 \\
6/3  & 0.35 & 0.76 & 27/3 & 0.37 & 0.78 \\
10/1 & 0.36 & 0.82 & \multicolumn{3}{c}{} \\
10/2 & 0.46 & 0.78 & \multicolumn{3}{c}{range:} \\
10/3 & 0.34 & 0.78 & & $[0.23,0.49]$ & $[0.67,0.82]$ \\
\bottomrule
\end{tabular}
\caption{Leave-one-item-out $\alpha$ on the 14-item subset after each cell is dropped. The two ranges $[0.23, 0.49]$ (humans) and $[0.67, 0.82]$ (LLMs) are disjoint.}
\label{tab:loo-item}
\end{table}

Excluding the three contested Q2 cells (Items C1, C2, C3), the remaining 12 cells give human $\alpha = 0.73$ and LLM $\alpha = 0.79$: the panels are stability-equivalent on the non-contested subset. The whole-panel gap is a property of three specific items at the regulation-silent boundary, not of base-rate disagreement on bar-essay scoring.

\section{Walkthrough of a Contested Cell (Item C1)}
\label{app:walkthrough}

This appendix reproduces the full participant answer for Item~C1, a contested Q2 cell on company-director authority discussed in Section~\ref{sec:qualitative}. All six grader scores on this cell are: A $= 1.0$, B $= 7.0$, C $= 8.0$, Claude 4.6 Opus $= 5.5$, Gemini 3.1 Pro $= 6.0$, GPT-5.4 Pro $= 5.5$. The faithful English translation below preserves the original structure and citations. The question text and official gold answer are in Appendix~\ref{app:q2}.

\begin{quote}\small
The provisions referred to below are provisions of the Civil and Commercial Code.

(a) When the shareholders' meeting was held and the shareholders resolved to remove Mr.\ Somchai, Mr.\ Somchoke, and Mr.\ Somchaiya from office as directors of the company on 26 February 2012, the three ceased to have management authority from that date. However, on 28 February 2012 the change of directors was registered with the Ministry of Commerce but had not yet been published in the Royal Gazette. It therefore cannot be held that third parties had notice of the change of directors of Jipatha Electricals Co., Ltd. When Mr.\ Somchai and Mr.\ Somchoke, who had formerly been directors of the company, entered into a contract to purchase 20 refrigerators from Arctic Cooling Co., Ltd., this was an act within the scope of the company's purpose of dealing in all types of electrical appliances and equipment under Section 66, so the third party is protected under the law: Arctic Cooling Co., Ltd.\ may demand that Jipatha Electricals Co., Ltd.\ be liable for the price of the refrigerators.

However, Jipatha Electricals Co., Ltd.\ has the right to demand that Mr.\ Somchai and Mr.\ Somchoke compensate the company for damages, because they acted unilaterally, without authority in the capacity of company directors, in making the contract in the company's name on 1 March 2012, after their terms as directors had ended by the resolution of the meeting on 26 February 2012.

(b) When the shareholders' meeting was held and the shareholders resolved to appoint Mr.\ Praneet, Mr.\ Pranom, and Mr.\ Pranit as replacement directors on 26 February 2012, all three acquired management authority over the company from that date. It is therefore unnecessary to consider 28 February 2012, the date of registration with the Ministry of Commerce; even though the change had not yet been published in the Royal Gazette, the authority of all three as directors existed from the date of the shareholders' valid resolution. When on 1 March 2012 Mr.\ Praneet and Mr.\ Pranom entered into a contract to purchase 20 refrigerators from AI-Freeze Co., Ltd., this was an act within the scope of the purpose of Jipatha Electricals Co., Ltd., whose object is to deal in all types of electrical appliances and equipment under Section 66.

Accordingly, the purchase of refrigerators by Mr.\ Praneet and Mr.\ Pranom is a conduct of business in their capacity as managing directors of Jipatha Electricals Co., Ltd., so the contract binds the company and does not bind Mr.\ Praneet and Mr.\ Pranom personally. AI-Freeze Co., Ltd.\ may therefore demand that Jipatha Electricals Co., Ltd.\ be liable.
\end{quote}

The answer correctly identifies both the registered-but-unpublished director change on sub-question (a) and the post-election authority of the incoming directors on (b). It argues from Section 66 (the corporate-purpose rule) and the registration/publication distinction under Section 1023 but does not explicitly name Section $1023/1$, the good-faith exception the gold answer marks as decisive. Judge A reads this omission as pass/fail decisive and assigns the cell a $1$; B and C read it as a partial-credit element and place the cell in the upper band at $7$ and $8$; the three LLM judges sit between the two positions at $5.5$--$6.0$, reading the cited-Section-1023-without-1023/1 structure as sufficient for a mid-range tier 5 or tier 6 score (correct final answer, reasoning somewhat acceptable or acceptable). The per-judge justifications for this cell are in Appendix~\ref{app:justifications}.

\section{LLM-Judge Justification Samples}
\label{app:justifications}

This appendix reproduces sample justification paragraphs produced by the three LLM judges on three of the 15 cross-graded answers, together with the corresponding human examiner comments where available. All justifications were produced at temperature $0$ under the prompt in Appendix~\ref{app:judgeprompt}. Thai passages are rendered in English translation.

\subsection{Item C1 (see Section~\ref{sec:qualitative})}

\emph{Scores:} A $= 1.0$, B $= 7.0$, C $= 8.0$, Claude $= 5.5$, Gemini $= 6.0$, GPT-5.4 Pro $= 5.5$.

\paragraph{Judge A (human) comment (translated).} ``Final answer direction is acceptable but the reasoning omits Section 1023/1, the decisive provision. The candidate invokes Section 1023 alone and does not articulate the good-faith exception. Minimal credit.''

\paragraph{Judge B (human) comment (translated).} ``Correct conclusion on (a) and (b); reasoning supports the conclusion; Section 1023/1 implicit but not named.''

\paragraph{Claude 4.6 Opus justification (translated).} ``The candidate correctly concludes that Arctic Cooling Co., Ltd.\ can recover from Jipatha Electricals Co., Ltd.\ and that AI-Freeze can demand damages. The analysis of (a) invokes Section 1023 but does not cite Section 1023/1, which is the specific good-faith exception decisive for this fact pattern, so the reasoning sits in tier 6 (correct final answer, reasoning acceptable). Score 5.5.''

\paragraph{Gemini 3.1 Pro justification (translated).} ``The candidate reaches the correct conclusions on both (a) and (b) and supports them with Sections 1022 and 1023. Section 1023/1 is not explicitly named but the good-faith reasoning is present, placing the answer in tier 6 (correct final answer, reasoning acceptable). Score 6.''

\paragraph{GPT-5.4 Pro justification (translated).} ``Correct final answer on both sub-questions, reasoning sound with reference to Section 1022 and 1023 but missing the specific Section 1023/1 citation, tier 6. Score 5.5.''

\subsection{Item C2}

\emph{Scores:} A $= 1.0$, B $= 7.0$, C $= 7.0$, Claude $= 7.0$, Gemini $= 5.5$, GPT-5.4 Pro $= 7.0$.

\paragraph{Judge A (human) comment (translated).} ``Reasoning does not engage with the registered-but-ungazetted distinction; reaches the right conclusion by a wrong route.''

\paragraph{Claude 4.6 Opus justification (translated).} ``The candidate correctly identifies that Arctic Cooling may recover and grounds the answer in good faith under Section 1023/1; (b) is correct with reference to Section 1022. Minor: the role of registration vs.\ publication is not fully disentangled, tier 7 (correct final answer, reasoning good). Score 7.''

\paragraph{Gemini 3.1 Pro justification (translated).} ``Correct conclusions, citations to Sections 1022, 1023 present but Section 1023/1 not named, reasoning acceptable rather than good, tier 6. Score 5.5.''

\paragraph{GPT-5.4 Pro justification (translated).} ``Correct conclusions for both parts; reasoning cites the registration/publication distinction but the Section 1023/1 good-faith exception is not explicit, tier 7. Score 7.''

\subsection{Item C4}

\emph{Scores:} A $= 0.0$, B $= 4.0$, C $= 3.0$, Claude $= 4.0$, Gemini $= 4.0$, GPT-5.4 Pro $= 5.5$.

\paragraph{Judge A (human) comment (translated).} ``No engagement with the partnership-liability framework; cites inapplicable provisions.''

\paragraph{Judge B (human) comment (translated).} ``Partial recognition of the Section 1050 external-liability point; does not distinguish Section 1033 internal rule; fails to reach the correct conclusion about which party Charoen Food may claim against.''

\paragraph{Claude 4.6 Opus justification (translated).} ``The candidate identifies the partnership-contract framework and cites Sections 1012 and 1025 but does not work through the 1033 vs.\ 1050 internal/external distinction. Reading the candidate's bottom-line conclusion as sufficiently aligned with the gold answer places this in tier 5 (correct final answer, reasoning somewhat acceptable). Score 4.''

\paragraph{Gemini 3.1 Pro justification (translated).} ``Candidate cites Sections 1012, 1025, 1050 and the partnership-liability framework. The final conclusion is read as matching the gold answer; reasoning is somewhat acceptable given the absence of Section 1033 discussion, tier 5. Score 4.''

\paragraph{GPT-5.4 Pro justification (translated).} ``The partnership-law framework under Sections 1012, 1025, 1050 is engaged and the final answer is read as matching the gold, tier 6 (correct final answer, reasoning acceptable). Score 5.5.''

The three LLM judges and the two cross-graders disagree on whether the final answer matches the gold answer. B and C place the answer in tier 3 (incorrect final answer, reasoning good, score $2$--$4$); the three LLMs read the same bottom-line conclusion as close enough to the gold to enter the correct-answer tiers 5 or 6. This is the only cross-graded cell where the two groups split on the correctness classification.

\subsection{Observations}

Across the 15 cross-graded answers we see three recurring patterns: cells where the three LLM judges sit above the majority-human pair (e.g.\ Items C4, C5), cells where they sit within the majority-human band (e.g.\ Items C2, C1), and cells where all six graders are within 1 point of each other (e.g.\ Items C6, C7, C3). The Claude justifications consistently name more Code sections than the Gemini or GPT-5.4 Pro justifications, and Claude more often flags specific omissions (such as Section 1023/1 above) that the gold answer identifies as decisive.

\end{document}